%% file: arxiv_publish.tex
\documentclass[10pt,twocolumn,letterpaper]{article}

\usepackage{iccv}              % To produce the CAMERA-READY version

\usepackage{colortbl}
\usepackage{array}
\usepackage{amsmath}
\usepackage{multirow}
\usepackage{makecell}
\usepackage{graphicx}
\usepackage{algorithmicx,algorithm}
\usepackage{color}
\usepackage{svg}
\usepackage{alltt}
\usepackage{makecell}

\usepackage{caption}
\usepackage{wrapfig}

\usepackage{algorithmicx,algorithm}
\usepackage[noend]{algpseudocode}
\usepackage{amsmath}
\usepackage{color}
\usepackage{diagbox}
\usepackage{amssymb}

\newcommand{\incre}[1]{\textcolor{teal!90}{#1}}

\definecolor{iccvblue}{rgb}{0.21,0.49,0.74}
\usepackage[pagebackref,breaklinks,colorlinks,allcolors=iccvblue]{hyperref}

\def\paperID{7578} % *** Enter the Paper ID here
\def\confName{ICCV}
\def\confYear{2025}

\title{Advancing Textual Prompt Learning with Anchored Attributes}

\author{
Zheng Li\textsuperscript{\rm 1},
Yibing Song\textsuperscript{\rm 2},
Ming-Ming Cheng\textsuperscript{\rm 1}, 
Xiang Li\textsuperscript{\rm 1}\thanks{Corresponding author.},
Jian Yang\textsuperscript{\rm 1}\footnotemark[1]\\
\textsuperscript{\rm 1} PCA Lab, VCIP, College of Computer Science, Nankai University, \\
\textsuperscript{\rm 2} DAMO Academy, Alibaba Group \\
{\tt\small zhengli97@mail.nankai.edu.cn, yibingsong.cv@gmail.com} \\
{\tt\small \{cmm, xiang.li.implus, csjyang\}@nankai.edu.cn} \\
}

\begin{document}
\maketitle

\begin{abstract}

Textual-based prompt learning methods primarily employ multiple learnable soft prompts and hard class tokens in a cascading manner as text inputs, aiming to align image and text~(category) spaces for downstream tasks. However, current training is restricted to aligning images with predefined known categories and cannot be associated with unknown categories.
In this work, we propose utilizing universal attributes as a bridge to enhance the alignment between images and unknown categories. Specifically, we introduce an \textbf{A}ttribute-anchored \textbf{T}extual \textbf{P}rompt learning method for vision-language models, named \textbf{ATPrompt}. This approach expands the learning space of soft prompts from the original one-dimensional category level into the multi-dimensional attribute level by incorporating multiple attribute tokens into the learnable soft prompts. Through this modification, we transform the text prompt from a category-centric form to an attribute-category hybrid form.
Additionally, we introduce a straightforward differentiable attribute search method to identify representative and suitable attributes for downstream tasks.
As an easy-to-use plug-in technique, ATPrompt can seamlessly replace the existing basic prompt format in textual-based methods, providing general improvements at a negligible computational cost. Extensive experiments across 11 datasets validate the effectiveness of our method. Code is publicly available at~\url{https://github.com/zhengli97/ATPrompt}. 

% Textual-based prompt learning methods primarily utilize multiple learnable prompt tokens and class tokens in a cascading manner, learning to map the representation of downstream tasks as closely as possible with that of the original pre-training tasks of the vision-language model. However, the current learning is confined to a one-dimensional class space, neglecting the potential of other dimensions. In this work, we introduce an \textbf{A}ttribute-templated \textbf{T}extual \textbf{P}rompt learning method for vision-language models, named \textbf{ATPrompt}. This method expands the original one-dimensional class learning space into a multi-dimensional attribute-class joint space by incorporating multiple distinct universal attribute bases. 

% 现有的提示学习方法目标是通过学习的方式，学出来适合的文本提示与下游图像进行对齐，训练过程中无法建立与新类别之间的连接，削弱模型的零样本泛化性能。

\end{abstract}
\section{Introduction}
Vision-Language Models~(VLMs)~\cite{lu2019vilbert,tan2019lxmert,jia2021scaling,radford2021learning,ma2024clip,sun2024alpha,li2024cascadeclip,li2024densevlm}, such as CLIP~\cite{radford2021learning,chen2023vlp} and ALIGN~\cite{jia2021scaling}, have demonstrated exceptional performance in recent years. These models are trained with a contrastive loss to establish alignment between image and text~(category) space. Inspired by the success of NLP~\cite{lester2021power,li2021prefix}, prompt learning~\cite{jia2022visual,zhou2022learning,zhou2022conditional} has emerged as a parameter-efficient tool to adapt powerful VLMs to downstream tasks. Models with a few learnable soft prompt tokens can achieve performance parity with, or even outperform, fully fine-tuned ones~\cite{jia2022visual}. Depending on how the soft prompt tokens are applied, existing methods can be broadly classified into textual-based~\cite{zhou2022learning,zhou2022conditional,khattak2023maple,khattak2023self,li2024promptkd,wu2024cascade} and visual-based approaches~\cite{bahng2022exploring,bar2022visual,jia2022visual,kunananthaseelan2024lavip,zhang2024dept}. Among these, the textual-based method is the most fundamental and straightforward, comprising the majority.

\begin{figure}[t]
    \centering
    \includegraphics[width=0.81\linewidth]{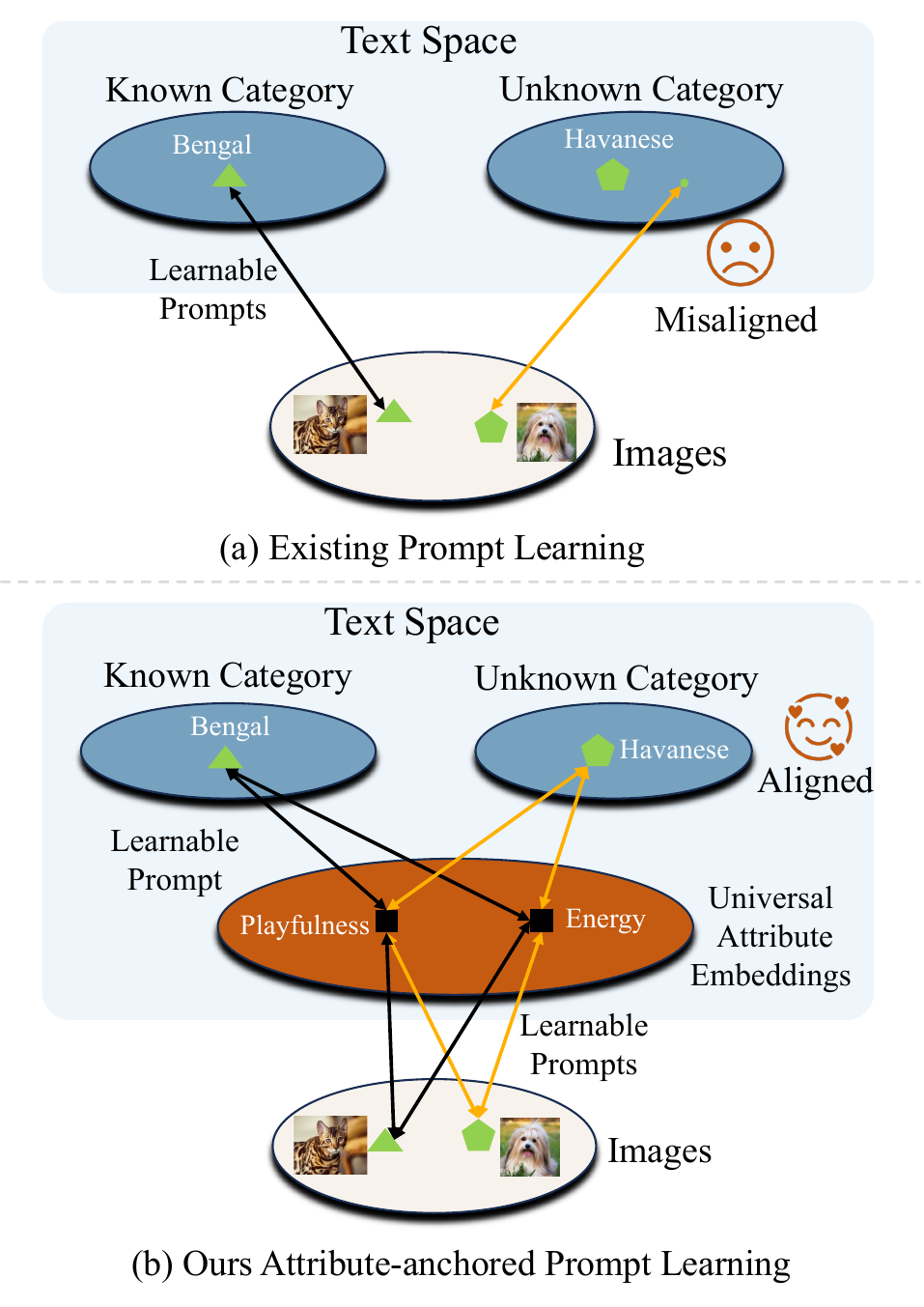}
    \vspace{-5pt}
    \caption{Comparison of image and text~(category) alignment processes through learnable prompts. (a) Current prompt learning methods align images with predefined categories but fail to establish accurate associations with unknown categories. (b) ATPrompt leverages universal attributes as an intermediary to create more accurate alignments between images and unknown categories.}
    \label{fig:attribute_space}
    \vspace{-15pt}
\end{figure}

In typical image classification tasks, current text-based methods~\cite{zhou2022learning,zhou2022conditional,khattak2023maple,khattak2023self} predominantly employ the traditional approach of concatenating learnable soft prompts with hard class tokens to replace handcrafted text prompts (e.g., “a photo of a \{class\}") as inputs to the encoder. Although this text prompt demonstrates strong performance, it restricts image alignment during training to predefined known categories only, thereby preventing accurate associations with unknown categories, as shown in Fig.~\ref{fig:attribute_space}(a). 
% Direct training can lead to overfitting on the known category training data and diminish the model's zero-shot generalization ability to unknown categories. 
Intuitively, when confronted with an unfamiliar category, humans often associate it with additional attributes~(e.g., color, shape or texture) to increase comprehensibility and clarity, rather than merely stating the object's name. For instance, one might describe a cheetah as: ``The cheetah is a cat-like animal with a \textit{small head}, \textit{short yellow hair}, and \textit{black spots}." or refer to an apple as ``That \textit{red spherical} fruit with \textit{orange stripes} is an apple." instead of using a general description such as ``This is a cheetah." or ``That fruit is an apple." \textit{Attributes can serve as bridges that connect unknown categories to our known knowledge.}
% The information provided by the additional attribute dimensions in the sentence can reduce the recognition complexity and improve the accuracy.

% , which expands the learning space from the original one-dimensional class level to a multi-dimensional attribute level by embedding multiple fixed attribute tokens into the learnable soft prompts. 
% Guided by the embedded attributes, we can build more accurate associations between images and unknown categories after training.
% \rewrite{Need rewrite.}
% Inspired by these observations, in this work, we present an \textbf{A}ttribute-embedded \textbf{T}extual \textbf{P}rompt (ATPrompt) learning method for Vision-Language Models (VLMs). By incorporating multiple universal attribute bases, our approach expands the learning space of soft prompt tokens from the original \textit{one-dimensional} category level to a \textit{multi-dimensional} attribute level. Aligning soft prompt tokens within the attribute space with the corresponding images facilitates a more accurate association between images and unknown classes, as illustrated in Fig.~\ref{fig:attribute_space}(b). Specifically, we embed multiple fixed universal attribute tokens into both the set of soft prompt tokens and class tokens, thereby transforming the original class-based textual prompt format into an attribute-class-mixed format.

Building on these observations, we propose a novel approach that leverages attributes as a bridge to enhance the alignment between images and unknown categories. Specifically, we introduce an attribute-anchored textual prompt learning method for VLMs, named ATPrompt. This method extends the learning space of soft prompts from the original one-dimensional category level to a multi-dimensional attribute level by integrating multiple fixed universal attribute tokens into the learnable soft prompts. Guided by these anchored attributes, soft tokens acquire not only category-specific but also attribute-related general representations during training. This results in improved alignment between images and unknown categories compared to the original method, as shown in Fig.~\ref{fig:attribute_space}(b). Additionally, based on the depth at which soft prompts are applied, we propose two versions of ATPrompt: shallow and deep, respectively, ensuring compatibility with existing methods of varying depths~\cite{zhou2022learning,khattak2023maple,khattak2023self}. To finalize these attributes, we present a simple and effective differentiable attribute search method that learns to identify suitable attributes from a candidate pool constructed by LLMs. The search operation only needs to be performed once per task, and once completed, the selected attributes can be used by ATPrompt for model training.

As an easy-to-use plug-in technique, ATPrompt can seamlessly substitute existing forms used in textual-based prompt learning methods, yielding general improvements with negligible additional computational overhead.

Our contributions can be summarized as follows:
\begin{itemize}
    \item We propose an efficient attribute-anchored textual prompt learning method that expands the learning space of soft prompts from a one-dimensional class level to the multi-dimensional attribute level.
    \item We introduce an effective differentiable search method to select appropriate attributes for downstream tasks.
    \item Both shallow and deep versions of ATPrompt are introduced to ensure compatibility with existing prompt learning methods of varying depths.
    \item Extensive experiments demonstrate that ATPrompt can be seamlessly integrated into existing textual-based methods, resulting in consistently improved performance with negligible computational costs.
\end{itemize}

\begin{figure*}[t]
    \centering
    \includegraphics[width=0.99\linewidth]{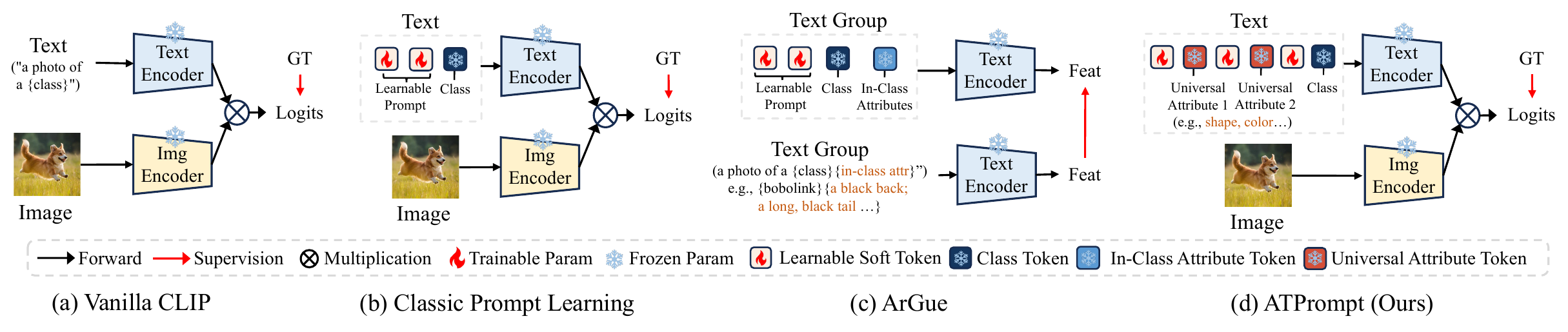}
    \vspace{-8pt}
    \caption{Architectural comparison among existing methods. (a) Vanilla CLIP employs a hand-crafted text template as input to the text encoder. (b) Classical prompt learning proposes a new text form that concatenates multiple learnable soft tokens with class tokens. (c) ArGue~\cite{tian2024argue} employs multiple \textit{in-class} attributes mined by LLMs as supplementary information. These attributes are utilized to construct distinct text groups, which serve as \textit{learning targets}, thereby regularizing the learning of soft tokens. The final prediction is achieved by ensembling all groups.
    (d) Our ATPrompt treats \textit{universal attributes} as \textit{learning components} and anchors them into existing soft prompt templates. Though this operation, we expand the learning space of soft tokens to multi-dimensional attribute levels and facilitate the alignment of images with unknown class texts.
    }
    \label{fig:framework}
    \vspace{-10pt}
\end{figure*}

\section{Related Work}
\textbf{Prompt Learning for VLMs.}
Inspired by recent advancements in NLP~\cite{lester2021power,li2021prefix}, prompt learning~\cite{zhou2022learning, zhou2022conditional, zhu2023prompt, khattak2023maple,roy2023consistency,khattak2023self,li2024promptkd,zhang2025comprompter} has garnered significant interest among vision researchers aiming to apply these techniques to VLMs~\cite{radford2021learning, jia2021scaling,yang2023clip}, such as CLIP.
% These approaches seek to adapt pre-trained VLMs to various downstream tasks by utilizing learnable soft prompts. 
% Serving as a parameter-efficient fine-tuning tool, prompt learning facilitates the adaptation of VLMs to new tasks. 
% Existing studies can be broadly classified into two categories: text-based~\cite{zhou2022learning,zhou2022conditional,zhu2023prompt,khattak2023maple} and image-based~\cite{jia2022visual,kunananthaseelan2023lavip,pei2024sa2vp} methods, with text-based methods being more prevalent. 
CoOp~\cite{zhou2022learning} is the pioneering text-based approach that introduced the concept of using a combination of soft textual tokens and a hard class token as input. 
% A class embedding can then be obtained through the text encoder for classification purposes. 
Subsequent studies \cite{zhou2022conditional,yao2023visual,khattak2023maple,lee2023read,khattak2023self,li2024promptkd,wu2025skip} have predominantly followed this textual prompt format. However, this form constrains the soft prompts to align with images within a one-dimensional, predefined category space, limiting their applicability to unknown categories. Therefore, training based on the current text form will be more likely to overfit to known categories, diminishing their zero-shot generalization capability for unknown categories. To address this limitation, several methods have been proposed~\cite{yao2023visual, kan2023knowledge, zhu2023prompt, khattak2023self, li2024promptkd}.
% To address this limitation and preserve generalization to unknown categories during training, several methods have been proposed~\cite{yao2023visual, kan2023knowledge, zhu2023prompt, khattak2023self, li2024promptkd}.
For instance, KgCoOp~\cite{yao2023visual} uses hand-crafted hard prompts to regularize learnable soft prompts during training. 
% ProGrad~\cite{zhu2023prompt} uses zero-shot prediction results to regularize the gradient during model training through KL loss. 
PromptSRC~\cite{khattak2023self} utilizes CLIP's~\cite{radford2021learning} original features to regularize the learning of soft prompts for image and text branches. 
PromptKD~\cite{li2024promptkd} utilizes a pre-trained strong teacher model to guide the learning of a student model~\cite{hinton2015distilling,li2023curriculum,yang2023online,li2024dual} with learnable prompts. 
Despite these advancements, none of the above methods address the inherent limitations of the format itself. 
In this work, we introduce the attribute-anchored textual prompt format for VLMs, which proposes to utilize attributes as a bridge to build more accurate associations between images and unknown categories.

\noindent\textbf{Attributes for VLMs.}
In practice, categories typically encompass multiple attributes. When individuals encounter an unfamiliar category, they often describe it using additional attributes to enhance the clarity of their communication, rather than merely stating its name. Inspired by this observation, numerous studies~\cite{menon2022visual, chen2023ovarnet, zhai2024multi, tian2024argue, wang2024learning} have begun to leverage attributes to support their objectives.
VCD~\cite{menon2022visual} was the pioneering work to propose the use of LLMs to decompose class names into multiple \textit{in-class} attributes~(e.g., beak and tail for birds) for classification. 
% HPT~\cite{wang2024learning} organizes numerous \textit{intra-class} attributes into a structured graph and designs a graph network to capture the relationships between entities and their attributes. 
AAPL~\cite{kim2024aapl} introduces a meta-network to extract visual attribute features based on encoded image features, facilitating image-text alignment. 
TAP~\cite{ding2024tree} presents a structured ``Tree of Attributes" approach that leverages attribute-specific knowledge graphs to enhance VLMs.
ArGue~\cite{tian2024argue} leverages large language models to mine multiple in-class attributes and integrates them into soft prompts and fixed templates to create multiple text groups. It employs the original text features generated by the fixed template to regularize the learning of soft tokens, as illustrated in Fig.~\ref{fig:framework}(c).
Most prior studies have focused on utilizing in-class attributes to enhance model performance by providing supplementary attribute information. However, when dealing with an unknown class, reacquiring the attributes of the new class becomes necessary—a process that is both complex and costly.
% In contrast, our work focuses on \textit{inter-class} attributes (e.g., color, shape, texture) and initially reformulates the textual prompt learning method~\cite{zhou2022learning} from a class-based prompt form to an attribute-class mixed form. Our approach can be seamlessly integrated into existing text-based methods, enhancing their performance without incurring additional computational costs.

In this work, we believe that universal~(inter-class) attributes are more efficient and robust than the in-class attributes used in previous works.
% focus on using LLMs to mine universal \textit{inter-class} attributes~(e.g., color, shape, texture) for prompt learning. 
Instead of taking attributes as the learning objective, we treat it as a learning component and propose to anchor universal attributes into soft prompt templates, transforming the existing class-centric form~\cite{zhou2022learning} into a hybrid attribute-class learnable textual prompt form. Our approach can be seamlessly integrated into existing textual-based methods, enhancing their performance without incurring additional computational costs.

% Existing works primarily focus on utilizing category descriptions or attributes obtained by simply querying LLMs with prompts such as ``What does a \{class\} look like?”. In contrast, we first let LLM summarize a pool containing multiple universal attributes. Inspired by DARTS~\cite{liu2018darts}, we further develop a differentiable attribute search method that learns to identify the attributes in the pool that best suit our method under the current downstream task, instead of using all attributes indiscriminately.

% \textbf{Differentiable NAS.}
% and utilize them to reformulate existing textual-based prompt learning approaches. 

% In this work, we first focus on the \textit{inter-class} common attributes and design an automatic pipeline that utilizes LLM to summarize these attributes in a \textit{multi-round dialogue} with feedback from the CLIP model.
% \noindent\textbf{Chain of Thoughts.}

\section{Method}

Prompt learning~\cite{zhou2022learning,zhou2022conditional,zhu2023prompt,khattak2023maple,khattak2023self} aims to enhance the generalization ability of pre-trained VLMs like CLIP on downstream tasks by training inserted learnable soft tokens. Existing textual-based methods all follow the classic prompt paradigm, concatenating soft prompt tokens and hard class tokens as the input to the text encoder, as shown in Fig.~\ref{fig:framework}(b). In this paper, we propose a simple and effective textual prompt learning method, named ATPrompt, which anchors multiple fixed universal attribute tokens into the original soft prompts, as shown in Fig.~\ref{fig:framework}(d).
% This operation expands the learning space of soft prompt tokens from the one-dimensional class level to the multi-dimensional attribute level. 
Guided by these attributes, soft prompts can learn not only category-specific but also attribute-related general representations through training. When encountering unknown categories, these learned attribute-related tokens can provide additional information to promote better image-text alignment.
% By aligning text prompts containing attribute embeddings with images during training, we can build more accurate associations with unknown categories through attributes.
% A detailed comparison with previous methods is illustrated in Fig.~\ref{fig:framework}.
Furthermore, to identify these universal attributes, we present an automated pipeline that encompasses sequential steps. Initially, we employ LLMs to synthesize the attribute pool for the current downstream category. Subsequently, we propose a differentiable attribute search method designed to identify attributes within the pool that are most suitable for our attribute-anchored prompt forms. For every task, this search operation is conducted \textit{only once}. Once the attributes are finalized, they are integrated into our ATPrompt for specialized model fine-tuning.

% In the following, we introduce the background of VLMs and prompt learning in Sec.~\ref{sec:pre}. Then we introduce our attribute-embedded prompt learning method in Sec.~\ref{sec:method} and the attribute search method in Sec.~\ref{sec:search}.
% Then we introduce our method in detail in Sec.~\ref{sec:method}.

\begin{figure*}[t]
    \centering
    \includegraphics[width=0.86\linewidth]{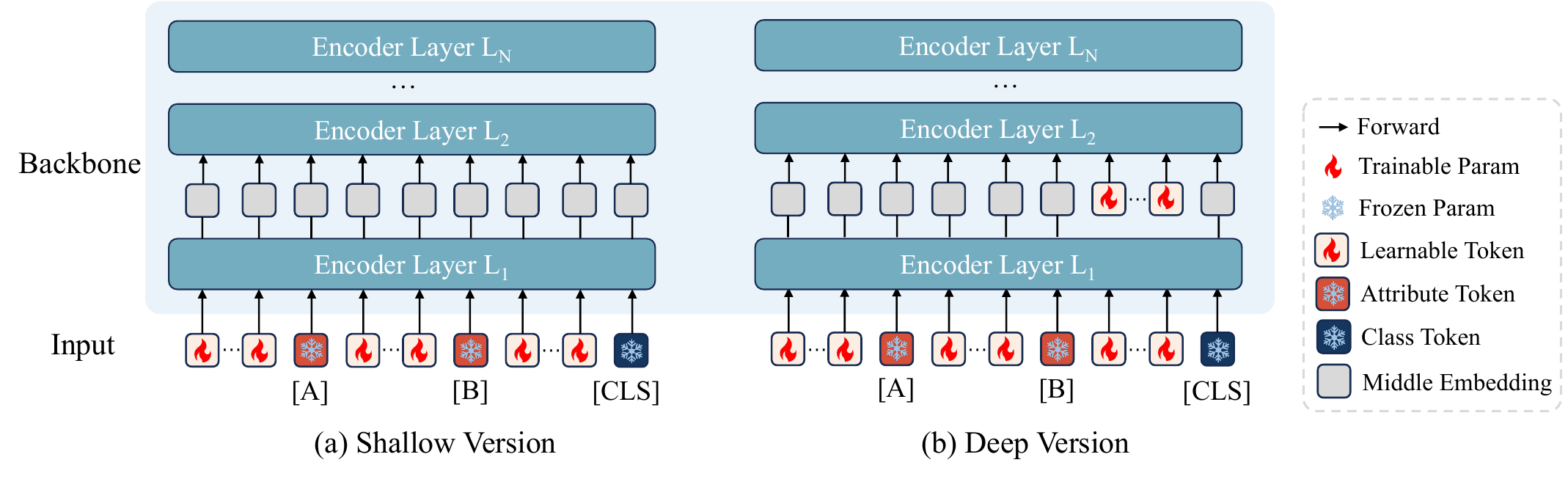}
    \vspace{-8pt}
    \caption{An illustration of the computation process for shallow and deep versions. Take two attributes $[\mathrm{A}]$ and $[\mathrm{B}]$ as examples. (a)~The shallow version concatenates hard attribute tokens, soft prompt tokens, and class tokens and inputs them into the encoder for calculation. 
    (b)~The deep version uses the same input but discards the class-related soft prompt tokens after calculating the self-attention and introduces them again before the next layer. These two forms can be compatible with existing methods of varying prompt depths, including input-level ones like CoOp~\cite{zhou2022learning}, CoCoOp~\cite{zhou2022conditional} and depth-level ones like MaPLe~\cite{khattak2023maple}.}
    \label{fig:shallow_and_deep}
    \vspace{-8pt}
\end{figure*}

\subsection{Preliminary}
\label{sec:pre}

\textbf{Vision-Language Models.}
Existing VLMs~\cite{radford2021learning,jia2021scaling}, such as CLIP, have demonstrated remarkable zero-shot generalization performance after training with 400 million image-text pairs. The primary objective of these models is to learn the alignment between image and text modalities produced by each encoder. Given a labeled image classification dataset $\mathrm{D}=\{(x, c)\}$ which includes $N$ class labels $C=\{c_{i}\}_{i=1}^{N}$, CLIP makes predictions by calculating the cosine similarity between image features and the text features of each class. Specifically, for each input image $x$, it undergoes feature extraction via the image encoder $h_{I}(x)$ and obtains a feature vector $u=h_{I}(x)$. Simultaneously, for each class, a series of textual descriptions $t$ are generated using the hand-crafted template. Then, these text descriptions are fed into the text encoder $h_{T}(x)$ to obtain text features $w=h_{T}(t)$. Finally, the output probability for image $x$ classified to $c$ is calculated as follows:
\begin{equation}
    p(c|x) = \frac{\exp(\cos(u, w_{c})/\tau)}{\sum_{i=1}^{N}\exp(\cos(u, w_{i})/\tau)}.
\label{equation:output_prob}
\end{equation}
where $\tau$ is the temperature parameter and $\cos(\cdot, \cdot)$ denotes cosine similarity.

\noindent\textbf{Prompt Learning for VLMs.}
Instead of manually designed hard prompts for image-text alignment, which is inaccurate and inflexible, recent prompt learning works~\cite{zhou2022learning,zhou2022conditional,yao2023visual,khattak2023maple,khattak2023self} like CoOp propose to learn appropriate soft textual prompts for downstream tasks. Concretely, $M$ learnable soft tokens $[T_{i}]_{i=1}^{M}$ are concatenated with the hard class token $[\mathrm{CLS}]$ as the input of the text encoder, as shown in Fig.~\ref{fig:framework}(b). Its form is shown as follows:
\begin{equation}
    P_{T}=[T_{1}][T_{2}]...[T_{M}][\mathrm{CLS}],
\label{equation:origin_prompt}
\end{equation}
where $M$ represents the length of soft tokens. For simplicity, we omit the prefix and suffix tokens in the input.

In addition to embedding soft tokens at the input level, existing studies~\cite{jia2022visual,khattak2023maple, khattak2023self, li2024promptkd} have also explored introducing them at deeper layers. This is achieved by adding soft tokens within the Transformer blocks and subsequently removing them after the self-attention computations. For the $i$-th block, this process can be described as follows:
\begin{equation}
     [\mathrm{CLS}_{i}] = L_{i}([\mathrm{T}_{i-1}, \mathrm{CLS}_{i-1}]).
\label{equtaion:deep1}
\end{equation}
where ${L_{i}}$ represents the $i$-th transformer block and $\mathrm{T}_{i}$ denotes the set of learnable soft tokens, defined as $\mathrm{T}_{i}=\{[T_{1}]_{i}, ..., [T_{M}]_{i}\}$.

\begin{figure*}[t]
    \centering
    \includegraphics[width=0.99\linewidth]{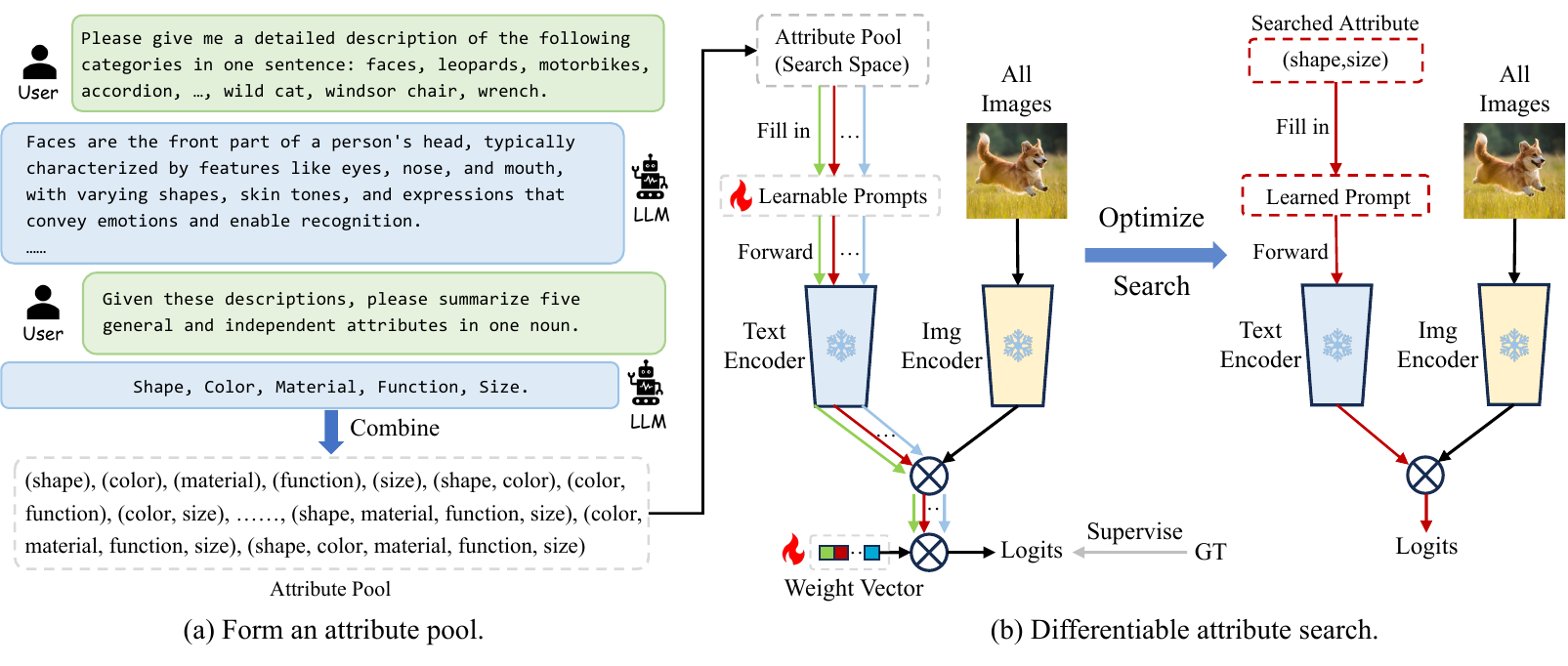}
    \vspace{-10pt}
    \caption{An overview of our attribute search pipeline. (a) We first query the LLM iteratively to obtain multiple independent attributes. These are subsequently aggregated to form a pool of candidate combinations, which serves as the input for the search process.
    (b) The forward computation for each candidate combination is represented by a distinct colored path. To identify the optimal attributes, we employ an alternating optimization algorithm that co-optimizes the soft tokens and a corresponding path weight vector. Upon completion of training, the combination associated with the highest-weighted path is selected as the final output.}
    \label{fig:attribute_search}
    \vspace{-8pt}
\end{figure*}

\subsection{Learning soft prompts with universal attributes}
\label{sec:method}

% In this paper, we propose expanding the original categorical space into a multi-dimensional attribute space by integrating hard attribute tokens within the soft prompt token set.
Our approach introduces two variants, distinguished by the number of layers at which soft tokens are applied: a shallow version and a deep version, as shown in Fig.~\ref{fig:shallow_and_deep}.

\noindent\textbf{Shallow Version.}
We begin by introducing the shallow version, in which hard attribute tokens are anchored solely at the input level, as illustrated in Fig.~\ref{fig:shallow_and_deep}(a). Consider two universal attributes, $\mathrm{A}$ and $\mathrm{B}$. According to Eqn.~\eqref{equation:origin_prompt}, the shallow-level text prompt $P_{T}$ provided to the text encoder can be expressed as follows:
\begin{equation}
    P_{T}=[T_{a_{1}}]...[T_{a_{m}}][\mathrm{A}][T_{b_{1}}]...[T_{b_{m}}][\mathrm{B}][T_{1}]...[T_{M}][\mathrm{CLS}].
\label{equation:class_end}
\end{equation}
where $a_{m}$ and $b_{m}$ are hyperparameters specifying the length of soft tokens for attributes A and B. In our method, we set these parameters to be the same by default.
% We usually 
% $[\mathrm{A}]$ and $[\mathrm{B}]$ denote the fixed hard attribution tokens that are not updated by loss backpropagation.

% Other than placing the hard class token at the end of the text prompt, we can also place it in the middle or front like:
% \begin{equation}
%     P_{mid}=[T_{a_{1}}]...[T_{a_{m}}][\mathrm{A}][T_{1}]...[T_{M}][\mathrm{CLS}][T_{b_{1}}]...[T_{b_{m}}][\mathrm{B}],
% \label{equation:class_middle}
% \end{equation}
% \begin{equation}
%     P_{front}=[T_{1}]...[T_{M}][\mathrm{CLS}][T_{a_{1}}]...[T_{a_{m}}][\mathrm{A}][T_{b_{1}}]...[T_{b_{m}}][\mathrm{B}].
% \label{equation:class_front}
% \end{equation}
% Other than placing the hard class token at the end of the text prompt, we can also place it in the middle or front.
% In Tab.~\ref{table:position}, we verify the performance of each position and select the best-performing end position as our default form.
While this example places the class token at the end, we also evaluate configurations where it is positioned at the front or in the middle. As shown in Tab.~\ref{table:position}, the end position yields the best performance and is therefore adopted as our default configuration.
% In the following experimental section, we verify the performance of different token positions. We choose the best-performing end position~(Eqn.~\eqref{equation:class_end}) as our default scheme.

\noindent\textbf{Deep Version.} 
In this version, learnable soft tokens are introduced at the input of the deep layers. Previous works, such as VPT and MaPLe, discard all soft tokens and subsequently reintroduce them after the block. When this operation is applied to attribute-related words, a gap will emerge between the introduced excessive low-level tokens and the existing high-level tokens, thereby weakening the feature continuity across layers. In this study, our approach selectively discards and then re-adds only class-related soft tokens in the input, specifically $[T_{1}], ..., [T_{M}]$, as shown in Fig.~\ref{fig:shallow_and_deep}(b). 
Based on Eqn.~\eqref{equtaion:deep1}, the deep version of ATPrompt can be rewritten as follows:
 \begin{equation}
    [\mathrm{F}_{1}, \_, \mathrm{CLS}_{1}] = L_{1}([\mathrm{T}_{a_{0}}, \mathrm{A}, \mathrm{T}_{b_{0}}, \mathrm{B}, \mathrm{T}_{0}, \mathrm{CLS}_{0}]),
\end{equation}
\begin{equation}
\begin{aligned}
    [\mathrm{F}_{i}, \_, \mathrm{CLS}_{i}] & = L_{i}([\mathrm{F}_{i-1}, \mathrm{T}_{i-1}, \mathrm{CLS}_{i-1}]). \\
    i & = 2,3,..., M.
\end{aligned}
\end{equation}
where $\mathrm{F}_{i}$ represents the features computed by the $i$-th Transformer layer. We demonstrate the effectiveness of this operation in Tab.~\ref{table:drop_operation}.

\noindent\textbf{Training.} Let $\theta$ represent the weight of the total soft tokens and let $v$ denote the selected fixed attribute tokens. Training is conducted on a labeled dataset $\mathrm{D}=\{(x, c)\}$, with the objective of minimizing the cross-entropy loss between predicted values and ground truth labels. This process can be formulated as follows:
\begin{equation}
    \underset{\theta}{\min}~ L_{train}= \underset{\theta}{\min}~\sum_{x \in D} \text{CE} (f(x; v, \theta), c).
\label{equation:attribute_prompt_training}
\end{equation}
where $f(\cdot)$ represents the function of the CLIP model. 

\subsection{Attribute search}
\label{sec:search}

% \rewrite{need rewrite}
% To finalize the attributes, two key aspects must be considered: selecting the appropriate content and determining the necessary quantity. A straightforward approach is to query the LLM directly. However, this method cannot adequately establish the optimal number of attributes for a specific downstream dataset. Additionally, without access to real data, querying solely by category name may introduce bias into the obtained results.
% To address these challenges, we propose an automated pipeline that can select the appropriate content and quantity for current downstream task, as illustrated in Fig.~\ref{fig:attribute_search}. 

Selecting the attributes involves two key considerations: their content and their quantity. While directly querying a LLM is a straightforward approach, it has notable drawbacks. This method cannot determine the optimal number of attributes for a specific downstream dataset, and querying by category name alone can introduce semantic bias. To address these issues, we propose an automated pipeline that selects the appropriate content and quantity of attributes for the current downstream task, as illustrated in Fig.~\ref{fig:attribute_search}.

\noindent\textbf{Attribute Pool.}
Inspired by CoT~\cite{wei2022chain, zhou2022least}, we divide the entire process into multiple steps to enhance the reasoning ability of LLMs. First, we prompt the LLM to generate descriptive sentences for each known category, thereby enriching category-related information. 
% Subsequently, 
% based on these sentences, 
% we prompt the LLM to summarize multiple independent attribute bases across these categories. 
Using these descriptions as context, we then prompt the LLM to summarize a set of independent attribute bases that are common across these categories.
An attribute pool is then formed by creating all possible combinations of these bases, as shown in Fig.~\ref{fig:attribute_search}(a). For $N$ attribute bases, this result in a total of $w=C_{N}^{1}+C_{N}^{2}+...+C_{N}^{N}$ candidates in the pool, which constitutes our search space.
Note that we do not consider the order of attributes, as permutations generally do not introduce significant semantic bias or affect the final performance, a claim we validate in our experiments.
% ($C^{1}_{5}+C^{2}_{5}+C^{3}_{5}+C^{4}_{5}+C^{5}_{5}$)
% \rewrite{we do not consider the order of attributes.}

\noindent\textbf{Attribute Searching.}
Inspired by DARTS~\cite{liu2018darts}, we introduce a differentiable attribute search method that learns to find representative attributes $v$ from the search space $\mathcal{V}$, as shown in Fig.~\ref{fig:attribute_search}(b). To make the search space continuous, we relax the discrete attribute selection into a softmax-weighted sum over all $w$ possible candidates:
\begin{equation}
f(x, v; \alpha, \theta) = \sum_{i\in \mathcal{V}}\frac{\exp(\alpha_{i})}{\sum_{i'\in \mathcal{V}}\exp(\alpha_{i'})} f(x, v_{i}; \theta).
\end{equation}
where $\alpha_{i}$ represents the weight for attribute combination $v_{i}$. The task of attribute search is thus reduced to learning the weight vector $\alpha$ for the candidate pool. 

After relaxation, our goal is to jointly learn the attribute weight $\alpha$ and soft prompt tokens $\theta$. Following standard practice~\cite{pham2018efficient,zoph2018learning,liu2018darts}, we optimize the weights~$\alpha$ by minimizing the validation loss $L_{val}$, while the soft tokens are learned by minimizing the training loss $L_{train}$. We employ an alternating algorithm~\cite{huang2020unfolding,li2023curriculum} to solve this bi-level optimization problem, alternating between these two subpromblems:

\begin{table*}[htb!]
    \begin{subtable}[t]{\textwidth}
    \centering
    \resizebox{0.85\linewidth}{!}
    {
    \centering
    \begin{tabular}{cccc|ccc|ccc|ccc}
    \toprule 
    % \hline
    \multirow{2}[3]{*}{Method} & \multicolumn{3}{c}{Average} & \multicolumn{3}{c}{ImageNet} & \multicolumn{3}{c}{ Caltech101} & \multicolumn{3}{c}{OxfordPets} \\
    \cmidrule(lr){2-4}\cmidrule(lr){5-7}\cmidrule(lr){8-10}\cmidrule(lr){11-13}
    & Base & Novel & HM & Base & Novel & HM & Base & Novel & HM & Base & Novel & HM \\
    \midrule
    % CLIP & 69.34 & 74.22 & 71.70 & 72.43 & 68.14 & 70.22 & 96.84 & {94.00} & 95.40 & 91.17 & 97.26 & 94.12\\
    CoOp~\scriptsize{\textcolor{gray}{(IJCV 22)}} & 82.69 & 63.22 & 71.66 & 76.47 & 67.88 & 71.92 & 98.00 & 89.81 & 93.73 & 93.67 & 95.29 & 94.47 \\
    CoCoOp~\scriptsize{(\textcolor{gray}{CVPR 22)}} & 80.47 & 71.69 & 75.83 & 75.98 & 70.43 & 73.10 & 97.96 & 93.81 & 95.84 & 95.20 & 97.69 & 96.43 \\
    MaPLe~\scriptsize{\textcolor{gray}{(CVPR 23)}} & 82.28 & 75.14 & 78.55 & 76.66 & 70.54 & 73.47 & 97.74 & 94.36 & 96.02 & 95.43 & 97.76 & 96.58 \\
    PromptSRC~\scriptsize{\textcolor{gray}{(ICCV 23)}} & 84.26 & 76.10 & 79.97 & 77.60 & 70.73 & 74.01 & 98.10 & 94.03 & 96.02 & 95.33 & 97.30 & 96.30 \\
    ArGue~\scriptsize{\textcolor{gray}{(CVPR 24)}} & 83.69 & 78.07 & 80.78 & 76.92 & 72.06 & 74.41 & 98.43 & 95.20 & 96.79 & 95.36 & 97.95 & 96.64 \\
    DePT~\scriptsize{\textcolor{gray}{(CVPR 24)}} & 83.66 & 71.82 & 77.29 & 77.13 & 70.10 & 73.45 & 98.33 & 94.33 & 96.29 & 94.70 & 97.63 & 96.14 \\
    CoPrompt~\scriptsize{\textcolor{gray}{(ICLR 24)}} & 84.00 & 77.23 & 80.48 & 77.67 & 71.27 & 74.33 & 98.27 & 94.90 & 96.55 & 95.67 & 98.10 & 96.87 \\
    PromptKD~\scriptsize{\textcolor{gray}{(CVPR 24)}} & 86.96 & 80.73 & 83.73 & 80.83 & 74.66 & 77.62 & 98.91 & 96.65 & 97.77 & 96.30 & 98.01 & 97.15 \\
    \midrule
    CoOp + ATPrompt & 82.68 & 68.04 & \textbf{74.65}~\scriptsize{\incre{(+2.99)}} & 76.27 & 70.60 & \textbf{73.33} & 97.95 & 93.63 & \textbf{95.74} & 94.77 & 96.59 & \textbf{95.67} \\ 
    % CoOp + ATPrompt & 82.68 & 68.04 & \textbf{74.65}~\scriptsize{\incre{(+2.99)}} & 76.27 & 70.70 & \textbf{73.33} & 97.95 & 93.63 & \textbf{95.74} & 94.77 & 96.59 & \textbf{95.67} \\ 
    CoCoOp + ATPrompt & 81.69 & 74.54 & \textbf{77.95}~\scriptsize{\incre{(+2.12)}} & 76.43 & 70.50 & \textbf{73.35} & 97.96 & 95.27 & \textbf{96.60} & 95.46 & 97.89 & \textbf{96.66} \\
    MaPLe + ATPrompt & 82.98 & 75.76 & \textbf{79.21}~\scriptsize{\incre{(+0.66)}} & 76.94 & 70.72 & \textbf{73.70} & 98.32 & 95.09 & \textbf{96.68} & 95.62 & 97.63 & \textbf{96.61} \\
    DePT + ATPrompt & 83.80 & 73.75 & \textbf{78.45}~\scriptsize{\incre{(+1.16)}} & 77.32 & 70.65 & \textbf{73.83} & 98.48 & 94.60 & \textbf{96.50} & 94.65 & 97.99 & \textbf{96.29} \\
    PromptKD + ATPrompt & 87.05 & 81.82 & \textbf{84.35}~\scriptsize{\incre{(+0.62)}} & 80.90 & 74.83 & \textbf{77.75} & 98.90 & 96.52 & 97.70 & 96.92 & 98.27 & \textbf{97.59} \\
    \bottomrule
    \end{tabular}
    }
    \end{subtable}

    \vspace{5pt}
    \begin{subtable}[t]{\textwidth}
    \centering
    \resizebox{0.82\linewidth}{!}
    {
    \begin{tabular}{cccc|ccc|ccc|ccc}
    \toprule \multirow{2}[3]{*}{ Method } & \multicolumn{3}{c}{StanfordCars} & \multicolumn{3}{c}{ Flowers102} & \multicolumn{3}{c}{ Food101} & \multicolumn{3}{c}{ FGVCAircraft} \\
    \cmidrule(lr){2-4}\cmidrule(lr){5-7}\cmidrule(lr){8-10}\cmidrule(lr){11-13} 
     & Base & Novel & HM & Base & Novel & HM & Base & Novel & HM & Base & Novel & HM \\
     \midrule
    % CLIP & 63.37 & 74.89 & 68.65 & 72.08 & 77.80 & 74.83 & 90.10 & 91.22 & 90.66 & 27.19 & 36.29 & 31.09 \\
    CoOp~\scriptsize{\textcolor{gray}{(IJCV 22)}} & 78.12 & 60.40 & 68.13 & 97.60 & 59.67 & 74.06 & 88.33 & 82.26 & 85.19 & 40.44 & 22.30 & 28.75 \\
    CoCoOp~\scriptsize{\textcolor{gray}{(CVPR 22)}} & 70.49 & 73.59 & 72.01 & 94.87 & 71.75 & 81.71 & 90.70 & 91.29 & 90.99 & 33.41 & 23.71 & 27.74 \\
    MaPLe~\scriptsize{\textcolor{gray}{(CVPR 23)}} & 72.94 & 74.00 & 73.47 & 95.92 & 72.46 & 82.56 & 90.71 & 92.05 & 91.38 & 37.44 & 35.61 & 36.50 \\
    PromptSRC~\scriptsize{\textcolor{gray}{(ICCV 23)}} & 78.27 & 74.97 & 76.58 & 98.07 & 76.50 & 85.95 & 90.67 & 91.53 & 91.10 & 42.73 & 37.87 & 40.15 \\
    ArGue~\scriptsize{\textcolor{gray}{(CVPR 24)}} & 75.64 & 73.38 & 74.49 & 98.34 & 75.41 & 85.36 & 92.33 & 91.96 & 92.14 & 40.46 & 38.03 & 39.21 \\
    DePT~\scriptsize{\textcolor{gray}{(CVPR 24)}} & 79.67 & 72.40 & 75.86 & 98.20 & 72.00 & 83.08 & 90.43 & 91.33 & 90.88 & 42.53 & 22.53 & 29.46 \\
    CoPrompt~\scriptsize{\textcolor{gray}{(ICLR 24)}} & 76.97 & 74.40 & 75.66 & 97.27 & 76.60 & 85.71 & 90.73 & 92.07 & 91.40 & 40.20 & 39.33 & 39.76 \\
    PromptKD~\scriptsize{\textcolor{gray}{(CVPR 24)}} & 82.80 & 83.37 & 83.13 & 99.42 & 82.62 & 90.24 & 92.43 & 93.68 & 93.05 & 49.12 & 41.81 & 45.17 \\
    \midrule
    CoOp + ATPrompt & 77.43 & 66.55 & \textbf{71.58} & 97.44 & 67.52 & \textbf{79.77} & 88.74 & 87.44 & \textbf{88.09} & 40.38 & 27.22 & \textbf{32.52} \\
    CoCoOp + ATPrompt & 74.50 & 73.47 & \textbf{73.98} & 96.52 & 73.59 & \textbf{83.51} & 90.59 & 91.74 & \textbf{91.16} & 37.30 & 33.15 & \textbf{35.10} \\
    MaPLe + ATPrompt & 75.39 & 73.84 & \textbf{74.61} & 97.82 & 75.07 & \textbf{84.95} & 90.65 & 92.00 & 91.32 & 37.61 & 36.15 & \textbf{36.87} \\
    DePT + ATPrompt  & 79.29 & 73.47 & \textbf{76.27} & 98.20 & 73.69 & \textbf{84.20} & 90.42 & 91.69 & \textbf{91.05} & 43.19 & 33.23 & \textbf{37.56} \\
    PromptKD + ATPrompt & 82.51 & 84.03 & \textbf{83.26} & 99.15 & 82.03 & 89.78 & 92.48 & 93.86 & \textbf{93.22} & 49.63 & 42.35 & \textbf{45.70} \\
    % PromptSRC & 78.27 & 74.97 &76.58 &98.07 &76.50 &85.95&90.67 &91.53 &91.10 &42.73 &37.87 & 40.15 \\
    % PromptKD  &80.48 & 81.78 & 81.12 & 98.69 & 81.91 &	89.52 & 89.43 & 91.27 & 90.34 &	43.61 & 39.68 & 41.55 \\
    % LLaMP & \textbf{81.56} & 74.54 & 77.89 & 97.82 & 77.40 & 86.42 & 91.05 & 91.93 & 91.49 & \textbf{47.30} & 37.61 & 41.90 \\
    % \midrule	
    % \rowcolor{mygray} 
    %  CoOp \emph{w}/TextRefiner & 71.40 & 70.90 & 71.15 & 95.92 & 74.33 & 83.76 & 90.88 & 91.43 & 91.15 &	35.35 & 35.87 & 35.61 \\
    %  \rowcolor{mygray} 
    %  PromptKD \emph{w}/TextRefiner & 80.91 & \textbf{81.83} & \textbf{81.37}& \textbf{99.30} & \textbf{82.91} & \textbf{90.37} & \textbf{91.42} &\textbf{92.71} &	\textbf{92.06} &	45.01	& \textbf{40.12}	& \textbf{42.42} \\
    \bottomrule
    \end{tabular}
    }
    \end{subtable}

    \vspace{5pt}
    \begin{subtable}[t]{\textwidth}
    \centering
    \resizebox{0.82\linewidth}{!}
    {
    \begin{tabular}{cccc|ccc|ccc|ccc}
    \toprule 
    \multirow{2}[3]{*}{Method} & \multicolumn{3}{c}{SUN397} & \multicolumn{3}{c}{DTD} & \multicolumn{3}{c}{EuroSAT} & \multicolumn{3}{c}{UCF101} \\
    \cmidrule(lr){2-4}\cmidrule(lr){5-7}\cmidrule(lr){8-10}\cmidrule(lr){11-13} 
    & Base & Novel & HM & Base & Novel & HM & Base & Novel & HM & Base & Novel & HM \\
    \midrule
    % CLIP  & 69.36 & 75.35 & 72.23 & 53.24 & 59.90 & 56.37 & 56.48 & 64.05 & 60.03 & 70.53 & 77.50 & 73.85 \\
    CoOp~\scriptsize{\textcolor{gray}{(IJCV 22)}} & 80.60 & 65.89 & 72.51 & 79.44 & 41.18 & 54.24 & 92.19 & 54.74 & 68.69 & 84.69 & 56.05 & 67.46 \\
    CoCoOp~\scriptsize{\textcolor{gray}{(CVPR 22)}} & 79.74 & 76.86 & 78.27 & 77.01 & 56.00 & 64.85 & 87.49 & 60.04 & 71.21 & 82.33 & 73.45 & 77.64 \\
    MaPLe~\scriptsize{\textcolor{gray}{(CVPR 23)}} & 80.82 & 78.70 & 79.75 & 80.36 & 59.18 & 68.16 & 94.07 & 73.23 & 82.35 & 83.00 & 78.66 & 80.77 \\
    PromptSRC~\scriptsize{\textcolor{gray}{(ICCV 23)}} & 82.67 & 78.47 & 80.52 & 83.37 & 62.97 & 71.75 & 92.90 & 73.90 & 82.32 & 87.10 & 78.80 & 82.74 \\
    ArGue~\scriptsize{\textcolor{gray}{(CVPR 24)}} & 81.52 & 80.74 & 81.13 & 81.60 & 66.55 & 73.31 & 94.43 & 88.24 & 91.23 & 85.56 & 79.29 & 82.31 \\
    DePT~\scriptsize{\textcolor{gray}{(CVPR 24)}} & 82.37 & 75.07 & 78.55 & 83.20 & 56.13 & 67.04 & 88.27 & 66.27 & 75.70 & 85.43 & 72.17 & 78.24 \\
    CoPrompt~\scriptsize{\textcolor{gray}{(ICLR 24)}} & 82.63 & 80.03 & 81.30 & 83.13 & 64.73 & 72.79 & 94.60 & 78.57 & 85.84 & 86.90 & 79.57 & 83.07 \\
    PromptKD~\scriptsize{\textcolor{gray}{(CVPR 24)}} & 83.69 & 81.54 & 82.60 & 85.84 & 71.37 & 77.94 & 97.54 & 82.08 & 89.14 & 89.71 & 82.27 & 86.10 \\
    \midrule
    CoOp + ATPrompt & 80.84 & 68.64 & \textbf{74.24} & 80.83 & 45.49 & \textbf{58.22} & 90.34 & 59.79 & \textbf{71.96} & 84.49 & 64.96 & \textbf{73.45} \\
    CoCoOp + ATPrompt & 80.50 & 76.86 & \textbf{78.64} & 78.63 & 56.89 & \textbf{66.02} & 87.95 & 74.15 & \textbf{80.46} & 82.74 & 76.40 & \textbf{79.44} \\
    MaPLe + ATPrompt & 80.98 & 78.15 & 79.54 & 80.50 & 58.28 & 67.61 & 94.84 & 77.59 & \textbf{85.35} & 84.08 & 78.88 & \textbf{81.40} \\
    DePT + ATPrompt & 82.42 & 76.48 & \textbf{79.34} & 82.64 & 56.77 & \textbf{67.30} & 89.60 & 69.50 & \textbf{78.28} & 85.60 & 73.15 & \textbf{78.89} \\
    PromptKD + ATPrompt & 83.87 & 81.35 & 82.59 & 86.92 & 72.34 & \textbf{78.96} & 97.05 & 92.07 & \textbf{94.49} & 89.29 & 82.44 & 85.73 \\
    % PromptSRC & 82.67 & 78.47 & 80.52 & 83.37 & 62.97 &71.75 & 92.90 & 73.90& 82.32 & 87.10 & 78.80 & 82.74 \\
    % PromptKD & 82.53 & \textbf{80.88} & 81.70 & 82.86 & 69.15 & 75.39 & 92.04 & 71.59 & 80.54 & 86.23 & 80.11 & 83.06 \\
    % LLaMP &\textbf{83.41}  & 79.90 & 81.62 & 83.49 & 64.49 & 72.77 &91.93  & \textbf{83.66} & \textbf{87.60} & 87.13 & 80.66 & 83.77 \\
    % \midrule
    % \rowcolor{mygray} 
    %  CoOp \emph{w}/TextRefiner &80.96	& 76.49	 &78.66 & 75.35 & 58.09 & 65.60 &	74.57 & 72.82 & 73.68 &	82.52 & 75.01 & 78.59 \\
    %  \rowcolor{mygray} 
    %  PromptKD \emph{w}/TextRefiner & 83.02 & 80.50 & \textbf{81.74} & \textbf{83.91} & \textbf{71.01} & \textbf{76.92} & 92.99 & 79.22 & 85.55& \textbf{89.20} & \textbf{81.90} & \textbf{85.39} \\		
    \bottomrule
    \end{tabular}
    }
    \end{subtable}
    \vspace{-5pt}
    \caption{Base-to-novel generalization experiments of five baselines with and without our ATPrompt on 11 datasets. Our method achieves consistent average performance improvement over different baselines.}
    \vspace{-5pt}
    \label{table:base_to_novel}
\end{table*}

\begin{table*}[ht]
    \centering
    \resizebox{0.87\linewidth}{!}
    {
        \begin{tabular}{ccccccccccccccc}
        \toprule
        \multirow{3}[3]{*}{Method} & Source & \multicolumn{10}{c}{Target Dataset} & \multirow{3}[3]{*}{Average} \\
        \cmidrule(lr){2-2}\cmidrule(lr){3-12}
         ~ & Image & Caltech & Oxford & Stanford & Flowers & \multirow{2}*{Food101} & FGVC & \multirow{2}*{SUN397} & \multirow{2}*{DTD} & Euro & \multirow{2}*{UCF101} &  \\
        ~ & Net    & 101     & Pets   & Cars      & 102     & ~ & Aircraft & ~ & ~ & SAT  & ~ & ~ \\
        \midrule
        CoOp & 71.51 & 93.70 & 89.14 & 64.51 & 68.71 & 85.30 & 18.47 & 64.15 & 41.92 & 46.39 & 66.55 & 63.88 \\
        +ATPrompt & 71.67 & 93.96 & 90.65 & 65.01 & 70.40 & 85.86 & 20.97 & 65.77 & 43.44 & 46.59 & 69.92 & \textbf{65.26} \scriptsize{\incre{(+1.38)}} \\
        \midrule
        CoCoOp & 71.02 & 94.43 & 90.14 & 65.32 & 71.88 & 86.06 & 22.94 & 67.36 & 45.73 & 45.37 & 68.21 & 65.74 \\
        +ATPrompt & 71.27 & 93.79 & 90.62 & 65.90 & 71.17 & 86.03 & 23.22 & 66.63 & 44.44 & 48.70 & 70.71 & \textbf{66.59} \scriptsize{\incre{(+0.85)}} \\
        \midrule
        MaPLe & 70.72 & 93.53 & 90.49 & 65.57 & 72.23 & 86.20 & 24.74 & 67.01 & 46.49 & 48.06 & 68.69 & 66.30 \\
        +ATPrompt & 70.69 & 94.04 & 91.03 & 66.06 & 71.99 & 86.33 & 24.42 & 67.05 & 45.21 & 48.63 & 69.15 & \textbf{66.75} \scriptsize{\incre{(+0.45)}} \\
        \bottomrule
        \end{tabular}
    }
    \vspace{-5pt}
    \caption{Cross-dataset generalization experiments of three baselines with and without our ATPrompt on 11 datasets. ATPrompt achieves consistent average performance improvement on target datasets. 
    % Our method achieves consistent average performance improvements over three baseline methods. 
    }
    \label{table:cross_dataset}
    \vspace{-5pt}
\end{table*}

% We apply the alternating algorithm~\cite{huang2020unfolding,li2023curriculum} to solve this problem, fixing one set of variables and solving for the other set. Formally, we can alternate between solving these two subproblems:
\begin{equation}
    \hat{\alpha}=\arg\underset{\alpha}{\min}~L_{val}(f(x, v; \alpha, \hat{\theta}), c),
\label{equation:find_attribute1}
\end{equation}
\begin{equation}
    \hat{\theta}=\arg\underset{\theta}{\min}~L_{train}(f(x, v; \hat{\alpha}, \theta), c).
\label{equation:find_attribute2}
\end{equation}
where $L_{train}$ and $L_{val}$ both use the cross-entropy loss function. After the search, the attribute combination with the highest weight ($\alpha_{i}$) is selected.
% At the end of the search, an attribute can be selected from the ones with the highest weights. 
% The search operation only needs to be performed once per task, and once completed, the selected attributes can be integrated into ATPrompt for targeted model training.
% After the search is complete, the selected attributes are integrated into ATPrompt for targeted model training. 

% \textbf{Overall.}
% The entire process comprises two steps. The first step is the attribute search stage. For downstream data, the attribute pool $\mathcal{V}$ is initially formed by LLMs. Subsequently, the attribute search method is conducted within this pool to identify the attribute $v$ that best suits our form. 

\noindent\textbf{Cost Analysis.}
Unlike traditional Neural Architecture Search~(NAS) methods \cite{liu2018progressive,elsken2019neural,tan2019mnasnet}, which search computationally expensive network-level parameters, our approach focuses on a lightweight token-level search space. This design makes our method significantly more efficient than previous approaches. In practice, our search converges in approximately 5 epochs, oftern requiring less than 5 minutes on a single A800 GPU for some datasets. In addition, search efficiency 
can be improved by curating a smaller set of base attributes, which narrows the search space.

\section{Experiments}
\label{section:exp}

\subsection{Settings}

% \textbf{Baseline Methods.} We apply our ATPrompt to a wide range of textual-based prompt learning approaches, including CoOp~\cite{zhou2022learning}, CoCoOp~\cite{zhou2022conditional}, KgCoOp~\cite{yao2023visual}, MaPLe~\cite{khattak2023maple}, and PromptSRC~\cite{khattak2023self}.

\noindent\textbf{Base-to-Novel Generalization.}
Following~\cite{zhou2022learning,zhou2022conditional,khattak2023self,li2024promptkd}, we split the dataset into base and novel classes. The model is trained on the base class training set and evaluated on the test set. 
The attributes applied in this experiment are searched based on the base class. 
For datasets without a dedicated validation set, like ImageNet, we split the 16-shot labeled data, using half for training and the other half for attribute search.

\noindent\textbf{Cross-dataset Experiments.}
Consistent with previous works~\cite{zhou2022learning,zhou2022conditional,khattak2023self}, we first train a model on the ImageNet-1K source dataset and then evaluate its generalization performance on several out-of-distribution datasets. The attributes used are obtained from the source dataset.

\begin{table}[t]
    \centering
    \resizebox{0.99\linewidth}{!}
    {
        \begin{tabular}{ccc}
        \toprule
        Dataset & Attribute Bases & Searched Results \\
        \midrule
        ImageNet & \makecell[c]{color, size, shape, \\ habitat, behavior} & (color, shape) \\
        \midrule
        Caltech101 & \makecell[c]{shape, color, material, \\ function, size} & (shape,size) \\
        \midrule
        OxfordPets & \makecell[c]{loyalty, affection, energy, \\ playfulness, intelligence} & (playfulness, energy) \\
        \midrule
        StanfordCars & \makecell[c]{design, engine, \\ performance,  luxury, color} & (luxury) \\
        \midrule
        Flowers102 & \makecell[c]{color, flower, \\ habitat, growth, season} & (color, habitat, growth) \\
        \midrule
        Food101    & \makecell[c]{flavor, texture, origin, \\ ingredients, preparation} & (flavor, preparation) \\
        \bottomrule
        \end{tabular}
    }
    \caption{Part of the results obtained after differentiable attribute search. Please refer to the appendix for the complete results.}
    \label{table:small_attribute_list}
    \vspace{-10pt}
\end{table}

\noindent\textbf{Attribute Search.} We select five independent attributes as the basis in the attribute pool. 
This results in 31 candidate attribute combinations for the search process. We use ChatGPT-4o for attribute queries. In Tab.~\ref{table:small_attribute_list} reports a subset of the queried attribute bases and the final combinations selected by our search. 
% The Appendix provides further details on our search process.

% We do not particularly consider the order of attributes, as the following experiments verify that it will not significantly affect the training results. 

\noindent\textbf{Implementation Details.}
% All the methods are implemented by PyTorch~\cite{paszke2019pytorch}. 
% We evaluate our method on 11 recognition datasets, including ImageNet-1K~\cite{deng2009imagenet}, Caltech-101~\cite{fei2004learning}, OxfordPets~\cite{parkhi2012cats}, StanfordCars~\cite{krause20133d}, Flowers-102~\cite{nilsback2008automated}, Food-101~\cite{bossard2014food}, FGVCAircraft~\cite{maji2013fine}, SUN-397~\cite{xiao2010sun},  DTD~\cite{cimpoi2014describing}, EuroSAT~\cite{helber2019eurosat} and UCF-101~\cite{soomro2012ucf101}. 
We evaluate the model performance on 15 datasets. We report base and novel class accuracy and their Harmonic Mean~(HM) averaged over 3 runs. The details of each dataset are attached in the Appendix.

\subsection{Base-to-Novel Generalization}

As demonstrated in Tab.~\ref{table:base_to_novel}, we evaluate the base-to-novel generalization performance of five baseline methods, both with and without integrating ATPrompt, across 11 diverse recognition datasets. Notably, ATPrompt consistently improves the average performance of all baseline methods.

\noindent\textbf{Reasons for Limited Improvement in Some Conditions.}
(1) Learnable text prompts are the central component in earlier works, and optimizing them through ATPrompt can obviously improve the performance. (2) Recent studies have expanded beyond learnable text prompts by introducing additional learnable modules. Since our work only involves the optimization of the learnable text prompt part, the improvement have become less obvious.

% (1) When we implement our method, we do not carefully adjust the hyperparameters of the baseline~(e.g., learning rate, training epochs, loss hyperparameters); they remain consistent with the original method. (2) Performance fluctuations due to limited training data.
% (1) In our experiments, we maintained the same loss hyperparameters as the original baseline method for fair comparison, without extensive tuning. (2) Deep methods like MaPLe and PromptKD typically involve many complex factors, which may make the improvement brought by our method less apparent in complex models.

\subsection{Cross-dataset Evaluation}
Tab.~\ref{table:cross_dataset} shows the cross-dataset generalization results for three baseline methods. Our method demonstrates superior performance, with improvements of $1.38\%$, $0.85\%$ and $0.45\%$ for CoOp, CoCoOp and MaPLe, respectively.

\subsection{Domain Generalization}
Tab.~\ref{table:domain_dataset} shows the domain generalization results for three baseline methods. The results show that our method improves CoOp, CoCoOp and MaPLe methods by $0.90\%$, $0.49\%$ and $0.33\%$ respectively.

\begin{table}[!h]
    \centering
    \resizebox{0.98\linewidth}{!}
    {
        \begin{tabular}{ccccccc}
        % \hline\noalign{\smallskip}
        \toprule
        \multirow{2}[2]{*}{Method} & Source & \multicolumn{4}{c}{Target Dataset} & \multirow{2}[2]{*}{Average} \\
        \cmidrule(lr){2-2}\cmidrule(lr){3-6}
         ~  & ImageNet & -V2 & -S & -A & -R & ~ \\
         \midrule
        % \hline\noalign{\smallskip}
        CoOp & 71.51 & 64.20 & 47.99 & 49.71 & 75.21 & 59.28 \\
        +ATPrompt & 71.67 & 64.43 & 49.13  & 50.91 & 76.24 & \textbf{60.18}~\scriptsize{\incre{(+0.90)}} \\
        % \hline\noalign{\smallskip}
        \midrule
        CoCoOp & 71.02 & 64.07 & 48.75 & 50.63 & 76.18 & 59.91 \\
        +ATPrompt & 71.27 & 64.66 & 49.15 & 51.44 & 76.33 & \textbf{60.40}~\scriptsize{\incre{(+0.49)}} \\
        % \hline\noalign{\smallskip}
        \midrule
        MaPLe  & 70.72 & 64.07 & 49.15 & 50.90 & 76.98 & 60.27 \\
        +ATPrompt & 70.69 & 64.40 & 49.10 & 51.77 & 77.11 & \textbf{60.60}~\scriptsize{\incre{(+0.33)}} \\
        % \hline
        \bottomrule
        \end{tabular}
    }
    % \vspace{-5pt}
    \caption{Domain generalization experiments of three baselines with and without our ATPrompt on four datasets. The integration of ATPrompt resulted in better generalization performance.
    }
    \vspace{-5pt}
    \label{table:domain_dataset}
\end{table}

\subsection{Further Analysis}
By default, the experiments are conducted on the ImageNet. To minimize the influence of other components, we mainly adopt CoOp as the baseline method. Two attributes~(color and shape) are used in our ATPrompt.
% For more experimental results, please refer to the Appendix.

\begin{figure}[h]
    \centering
    \includegraphics[width=0.62\linewidth]{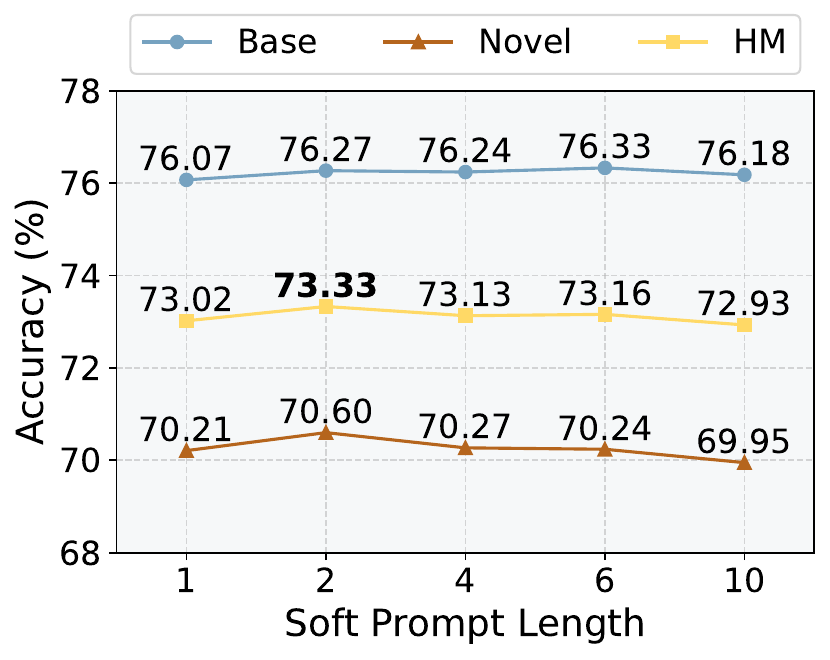}
    \vspace{-7pt}
    \caption{Illustration of varying soft token length on ImageNet. Increased tokens can lead to overfitting to the base class and weaken generalization to the novel class.}
    \label{fig:prompt_length}
    \vspace{-5pt}
\end{figure}

\noindent\textbf{Soft Prompt Length.} In Fig.~\ref{fig:prompt_length}, we examine the optimal soft token length for both attribute and class tokens. By varying the length from 1 to 10, we observe that longer prompts dilute the guiding influence of the attribute tokens, thereby reducing generalization to novel classes.

% \begin{table}[h]
%     \centering
%     \resizebox{0.54\linewidth}{!}{
%     \begin{tabular}{ccccc}
%         \hline\noalign{\smallskip}
%         Length & Base & Novel & HM \\
%         \hline\noalign{\smallskip}
%         1 & 76.07 & 70.21 & 73.02 \\
%         2 & 76.27 & \textbf{70.60} & \textbf{73.33} \\
%         4 & 76.24 & 70.27 & 73.13 \\
%         6 & \textbf{76.33} & 70.24 & 73.16 \\
%         10 & 76.18 & 69.95 & 72.93 \\
%         \hline
%     \end{tabular}
%     }
%     \caption{Comparison of varying prompt length on ImageNet. Increased prompt length can lead to overfitting to the base class and weaken generalization to the novel class.}
%     \label{table:prompt_length}
%     \vspace{-10pt}
% \end{table}

\noindent\textbf{Class Token Position.} 
The relative positioning of attribute tokens and class tokens in our method requires careful consideration. In Tab.~\ref{table:position}, we examine configurations where the class token is positioned in the middle or on either side of two attribute tokens.
% corresponding to Eqn.~\eqref{equation:class_middle}, \eqref{equation:class_front}, and \eqref{equation:class_end}, respectively. 
The results demonstrate that optimal performance is achieved when the class token is placed at the end, which aligns with the findings of CoOp.

\begin{table}[ht]
    \centering
    \resizebox{0.56\linewidth}{!}
    {
        \begin{tabular}{cccc}
        \toprule
        Position & Base  & Novel & HM    \\
        \midrule
        Front  & 76.12 & 70.50 & 73.20 \\
        Middle & 76.13 & 70.29 & 73.09 \\
        End    & \textbf{76.27} & \textbf{70.60} & \textbf{73.33} \\
        \bottomrule
        \end{tabular}
    }
    % \vspace{-5pt}
    \caption{Comparison of different class token positions on ImageNet. The end position works best.}
    \label{table:position}
    \vspace{-5pt}
\end{table}

% In contrast, previous studies such as VPT and MaPLe eliminate all soft tokens and subsequently reintroduce them in deeper layers.

\noindent\textbf{Prompt Operation of Deep Version.} In ATPrompt-Deep, we exclusively drop class soft tokens while retaining both hard and soft attribute tokens after they pass through the block. In this part, we compare the performance of partial drop~(i.e., removing attribute soft tokens while retaining hard tokens) and full drop~(i.e., removing both attribute soft and hard tokens) operations, as illustrated in Tab.~\ref{table:drop_operation}.
% A detailed structural comparison diagram is provided in the Appendix.

\begin{table}[h]
    \centering
    \resizebox{0.85\linewidth}{!}
    {
        \begin{tabular}{cccc}
        \toprule
        Operation of Attribute Token & Base  & Novel & HM    \\
        \midrule
        Retain all hard and soft tokens & \textbf{76.94} & \textbf{70.72} & \textbf{73.70} \\
        Partial drop and re-add & 76.87 & 70.44 & 73.51 \\
        Full drop and re-add    & 76.83 & 70.10 & 73.31 \\
        % 把attribute跟着一起引入到depth里面，并且随着learnable token一起drop
        \bottomrule
        \end{tabular}
    }
    % \vspace{-5pt}
    \caption{Comparison of operations on deep soft and hard attribute tokens based on MaPLe+ATPrompt. Preserving hard and soft attribute tokens in deep layers performs better than other operations.}
    \label{table:drop_operation}
    \vspace{-5pt}
\end{table}

% The results demonstrate that the drop and re-add operations lead to performance degradation, particularly with the full drop operation. 
The results show that maintaining the attributes of both hard and soft tokens during the forward process results in optimal performance. Conversely, dropping and reintroducing either hard or soft attribute tokens harms performance, likely because it disrupts the continuity of attribute representations across layers and complicates optimization.

% the actions of discarding and reintroducing attribute tokens diminish the model's effectiveness. A possible explanation for this is that such operations disrupt the continuity of attribute representations across layers and amplify the disparity with existing tokens, thereby complicating the training of soft tokens.

% For example, phrases like ``a yellow round leaf" and ``a round yellow leaf" convey the same meaning. 

\noindent\textbf{Attribute Order.}
In this study, we do not specifically focus on the order of attributes because varying the sequence usually does not result in semantic deviations in reality. 
Tab.~\ref{table:order_of_attribute} quantitatively assesses the impact of attribute order on prompt learning performance. From this table, we observe that despite variations in order, similar results are consistently produced, and the performance fluctuations across different orders remain within a reasonable range.

\begin{table}[ht]
    \centering
    \resizebox{0.59\linewidth}{!}
    {
        \begin{tabular}{cccc}
        \toprule
        Attributes     & Base  & Novel & HM    \\
        \midrule
        (shape, color) & 76.32 & 70.39 & 73.24 \\
        (color, shape) & 76.27 & 70.60 & \textbf{73.33} \\
        \midrule
        (size, habitat) & 76.44 & 70.23 & 73.20 \\
        (habitat, size) & 76.46 & 70.16 & 73.14 \\
        \bottomrule
        \end{tabular}
    }
    % \vspace{-5pt}
    \caption{Comparison of different orders on ImageNet. The order of attributes does not significantly affect the model, and performance fluctuations are within a reasonable range.}
    \label{table:order_of_attribute}
    \vspace{-5pt}
\end{table}

\noindent\textbf{Comparison to Other Attributes.} In Tab.~\ref{table:irrelevant_attribute}, we explore the effectiveness of attributes derived through alternative methods, specifically by manually selecting class-irrelevant and common attributes. 
It shows that manually selected irrelevant attributes exhibit comparable performance during training; however, they perform poorly when applied to new categories. This suggests that incorrect attribute tokens cause the soft tokens to develop biased representations, thereby diminishing their zero-shot generalization ability.

\begin{table}[h]
    \centering
    \resizebox{0.80\linewidth}{!}
    {
        \begin{tabular}{ccccc}
        % \hline\noalign{\smallskip}
        \toprule
        Type & Attributes & Base  & Novel & HM    \\
        % \hline\noalign{\smallskip}
        % Baseline &          & 88.33 & 82.26 & 85.19 \\
        % \hline\noalign{\smallskip}
        \midrule
        \multirow{2}*{Common} & (shape, size)    & 82.83 & 67.13 & 74.16 \\
        ~                     & (color, texture) & 82.73 & 67.56 & 74.38 \\
        % \hline\noalign{\smallskip}
        \midrule
        \multirow{2}*{Irrelevant} & (plane, engines) & 82.81 & 66.22 & 73.59 \\
        ~          & (football, sport) & 82.77 & 67.14 & 74.14 \\
        % \hline\noalign{\smallskip}
        \midrule
        Searched & - & 82.68 & \textbf{68.04} & \textbf{74.65} \\
        \bottomrule
        \end{tabular}
    }
    % \vspace{-5pt}
    \caption{Comparison of average performance for various attribute configurations on 11 datasets. The attributes obtained by our method achieve the best performance.}
    \label{table:irrelevant_attribute}
    \vspace{-5pt}
\end{table}

\section{Conclusion}

In this work, we introduce ATPrompt, an attribute-anchored textual prompt learning method that uses universal attributes as a bridge to improve generalization from seen to unseen categories. Our approach expands the learning space of soft prompts from a one-dimensional, category-centric structure to a multi-dimensional attribute space by anchoring fixed attribute tokens with the prompt. To ensure the selection of optimal attributes, we propose an automated pipeline designed to identify the most suitable candidates for any given downstream task. ATPrompt is designed with both shallow and deep architectural variants, rendering it broadly compatible with existing prompt learning methods. Extensive experiments validate the effectiveness of our approach. We believe this work offers a new direction for research into the fundamental structure of learnable prompts in prompt learning area.

\noindent\textbf{Limitations and future works.} 
This work is a preliminary study of the basic prompt form, which has not been able to achieve comparable performance to regularization-based methods when working alone. Furthermore, the current selection of attribute anchors relies on manual experimentation, and automatically discovering the optimal anchor positions through a learning-based approach remains a promising direction for future research.

\section{Acknowledgments} This work was supported by the National Natural Science Foundation of China under Grant Nos. 62361166670 and U24A20330. This research was also supported by the Young Scientists Fund of the National Natural Science Foundation of China~(Grant No.62206134), the Fundamental Research Funds for the Central Universities 070-63253222 and the Tianjin Key Laboratory of Visual Computing and Intelligent Perception~(VCIP). Computation is supported by the Supercomputing Center of Nankai University~(NKSC).

{
    \small
    \bibliographystyle{ieeenat_fullname}
    \bibliography{main}
}

\newpage
\maketitlesupplementary

\setcounter{table}{0}
\setcounter{figure}{0}
\setcounter{section}{0}
\setcounter{equation}{0}

\renewcommand{\thesection}{S\arabic{section}}
\renewcommand{\thetable}{S\arabic{table}}
\renewcommand{\thefigure}{S\arabic{figure}}
\renewcommand{\theequation}{S\arabic{equation}}

\section{Implementation Details}

\subsection{Dataset}
We evaluate the performance of our method on 15 recognition datasets. For generalization from base-to-novel classes and cross-dataset evaluation, we evaluate the performance of our method on 11 diverse recognition datasets. Specifically, these datasets include ImageNet-1K~\cite{deng2009imagenet} and Caltech-101~\cite{fei2004learning} for generic object classification; OxfordPets~\cite{parkhi2012cats}, StanfordCars~\cite{krause20133d}, Flowers-102~\cite{nilsback2008automated}, Food-101~\cite{bossard2014food}, and FGVCAircraft~\cite{maji2013fine} for fine-grained classification, SUN-397~\cite{xiao2010sun} for scene recognition, UCF-101~\cite{soomro2012ucf101} for action recognition, DTD~\cite{cimpoi2014describing} for texture classification, and EuroSAT~\cite{helber2019eurosat} for satellite imagery recognition. For domain generalization experiments, we use ImageNet-1K~\cite{deng2009imagenet} as the source dataset and its four variants as target datasets including ImageNet-V2~\cite{recht2019imagenet}, ImageNet-Sketch~\cite{wang2019learning}, ImageNet-A~\cite{hendrycks2021natural}, and ImageNet-R~\cite{hendrycks2021many}. 

\subsection{Attribute Search}
Inspired by DARTS~\cite{liu2018darts}, we employ a differentiable search method to identify the optimal content and quantity of attributes for our proposed attribute-anchored form. The search process is conducted for 10 epochs with a batch size of 32. We use SGD to optimize the soft prompts $\theta$ with an initial learning rate of $0.002$. and Adam to optimize the weight vector $\alpha$ with an initial learning rate of $0.02$. In our experiments, we use 5 attribute bases, which generate 31~(i.e., $C_{5}^{1}+C_{5}^{2}+C_{5}^{3}+C_{5}^{4}+C_{5}^{5}$) candidate combinations for the search process.

Tab.~\ref{table:attribute_list} presents the five attribute bases generated by the LLM, alongside the optimal attribute combination identified after the search. Furthermore,  Tab.~\ref{table:combination} displays the final weights of all candidate combinations from the search stage on the Caltech-101 dataset.

\subsection{Base-to-Novel Generalization}

\textbf{Baseline Methods.} To evaluate ATPrompt, we integrate it with several leading textual-based prompt learning approaches, including CoOp~\cite{zhou2022learning}, CoCoOp~\cite{zhou2022conditional}, MaPLe~\cite{khattak2023maple}, DePT~\cite{zhang2024dept} and PromptKD~\cite{li2024promptkd}. The experimental settings are detailed below.

\begin{figure*}[t]
    \centering
    \includegraphics[width=0.75\linewidth]{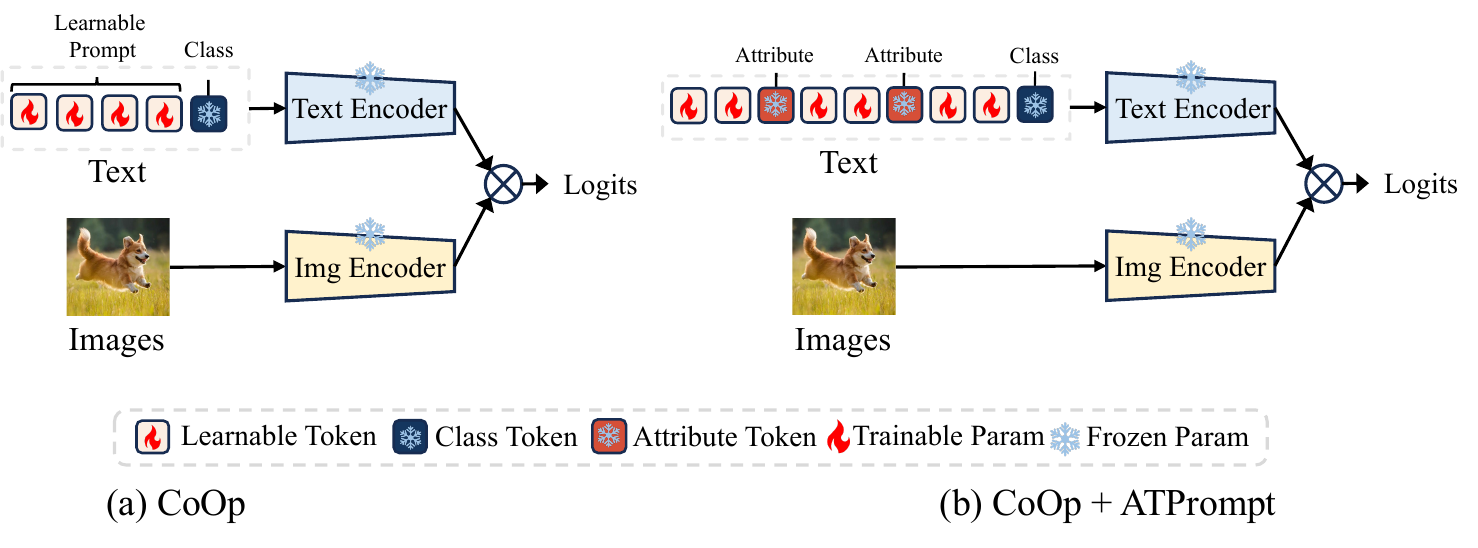}
    \vspace{-10pt}
    \caption{Architectural comparison between CoOp and CoOp+ATPrompt.}
    \label{fig:comparison_coop}
\end{figure*}

\noindent\textbf{Settings.} Our framework is implemented in PyTorch~\cite{paszke2019pytorch} and all experiments were conducted on a single NVIDIA A800 GPU. Following the baseline methods, we use a standard data augmentation scheme of random resized cropping and flipping. We employe Stochastic Gradient Descent~(SGD) as the optimizer. By default, the soft token lengths for attribute and class tokens are set to be identical, as attribute and class tokens are considered equally important. 
The specific implementation details for each baseline method are presented as follows:

\noindent\textbf{CoOp+ATPrompt:} Following the baseline, we use a batch size of 32 and an initial learning rate of 0.002. The original paper reports a learnable prompt length of $M=16$ for ResNet-50 but does not specify a length for ViT-B/16. In our setup, we set the sofo token length for both the attribute and class tokens to 2. While the baseline model is trained for 200 epochs, we reduce the training to 100 epochs while maintaining the same cosine decay schedule. Figure~\ref{fig:comparison_coop} illustrates the architectural differences between the original CoOp and CoOp+ATPrompt.

\noindent\textbf{CoCoOp+ATPrompt}: We adhere to the baseline's settings with a batch size of 1 and an initial learning rate of 0.002. Whereas the original paper specifies a soft class token length of 4, we set the length of our learnable tokens for the attribute and class token to 2. We adopt the same training schedule as the baseline: 10 epochs with cosine decay.

% In line with the baseline method, we employ a batch size of 1 and an initial learning rate of 0.002. The original paper specifies a learnable class prompt length of 4. In contrast, our method sets the lengths of both the learnable attribute soft prompts, $a_{m}$ and $b_{m}$, as well as the learnable class soft prompt, $M$, to 2. We follow the same training schedule as the baseline method, which involves training the model over 10 epochs using cosine decay.

\begin{figure*}[ht!]
    \centering
    \includegraphics[width=0.95\linewidth]{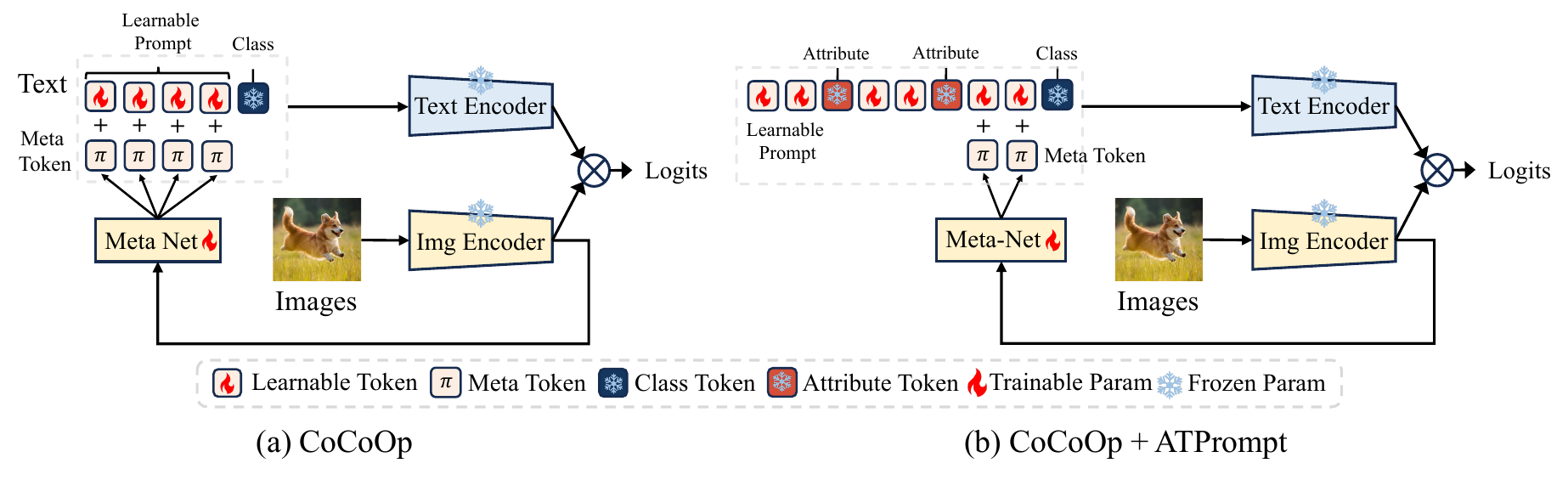}
    \vspace{-10pt}
    \caption{Architectural comparison between CoCoOp and CoCoOp+ATPrompt. In CoCoOp+ATPrompt, meta tokens are only added as offsets to class soft tokens.}
    \label{fig:comparison_cocoop}
\end{figure*}

CoCoOp's original design uses a meta-network to generate offsets for all soft prompt tokens. We retain this meta-network but modify its application: the meta tokens now serve as offsets exclusively for the class soft tokens, $[T_{1}], ..., [T_{M}]$, as shown in Fig.~\ref{fig:comparison_cocoop}.

% In the original paper, CoCoOp introduces a meta network that processes image features to generate meta tokens, which serve as offsets for all soft prompt tokens. In our CoCoOp+ATPrompt, we maintain both the meta network and the meta tokens; however, the meta tokens are exclusively used as offsets for class soft tokens $[T_{1}], ..., [T_{M}]$, as illustrated in Fig.~\ref{fig:comparison_cocoop}.

\noindent\textbf{MaPLe+ATPrompt}:
We adhere to the baseline hyperparameters, utilizing a batch size of 4 and an initial learning rate of 0.0035. We diverge from the original prompt configuration; whereas the baseline sets the learnable prompt length to 2, our method sets the soft token lengths of both the attribute and the class token to 4. The training schedule remains consistent with the baseline.

\begin{figure*}[ht!]
    \centering
    \includegraphics[width=0.85\linewidth]{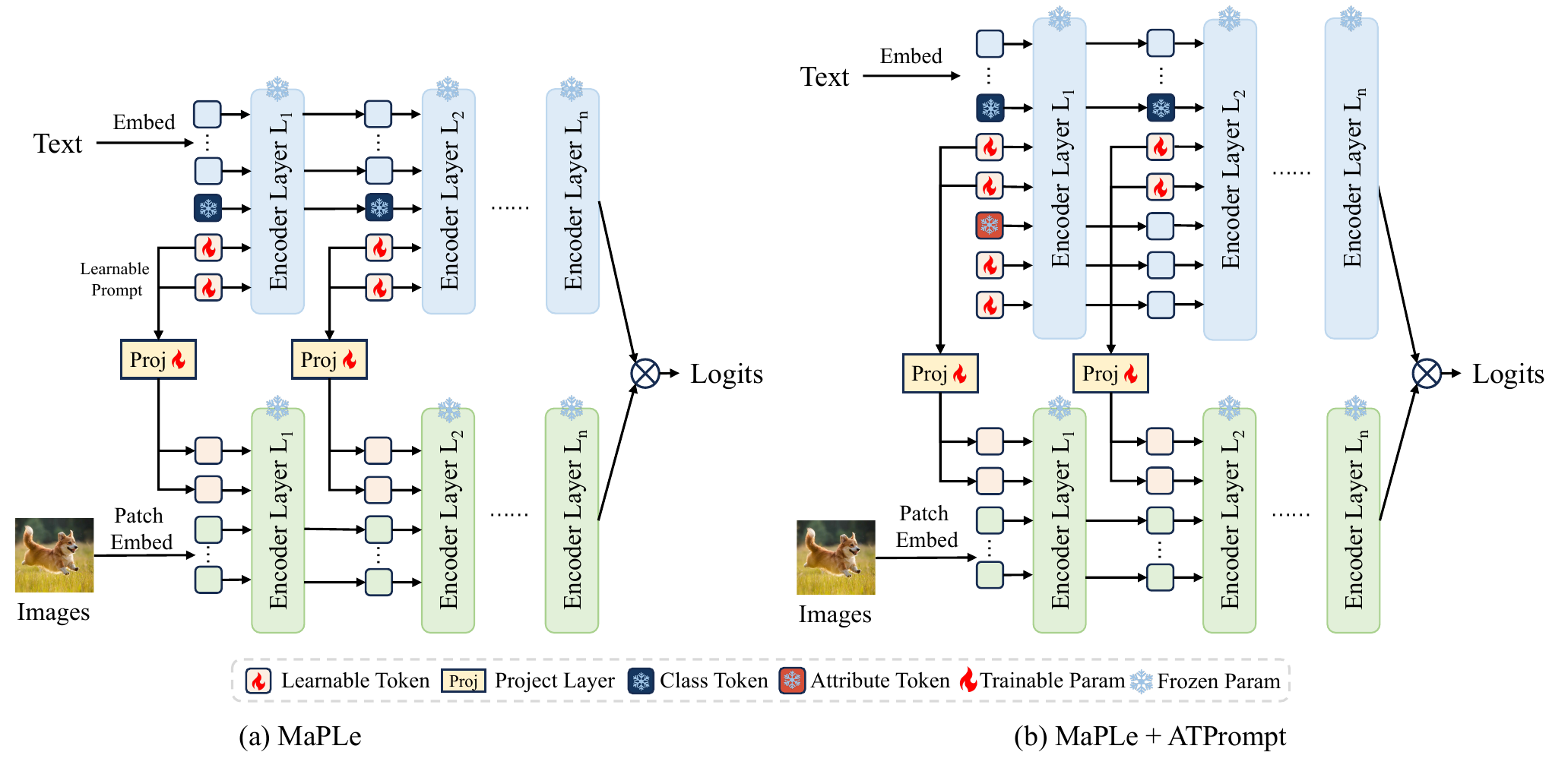}
    \vspace{-10pt}
    \caption{Architectural comparison between MaPLe and MaPLe+ATPrompt. }
    \label{fig:comparison_maple}
\end{figure*}

The primary architectural modification in MaPLe + ATPrompt concerns the projection mechanism. The original MaPLe framework inputs all textual soft tokens into a projection layer to generate corresponding visual tokens, which are then fused into the image encoder. Our approach, however, selectively inputs only the class soft tokens into this projection layer, while the attribute tokens are preserved without modification. This architectural difference is visualized in Fig.~\ref{fig:comparison_maple}.

\begin{figure*}[ht!]
    \centering
    \includegraphics[width=0.75\linewidth]{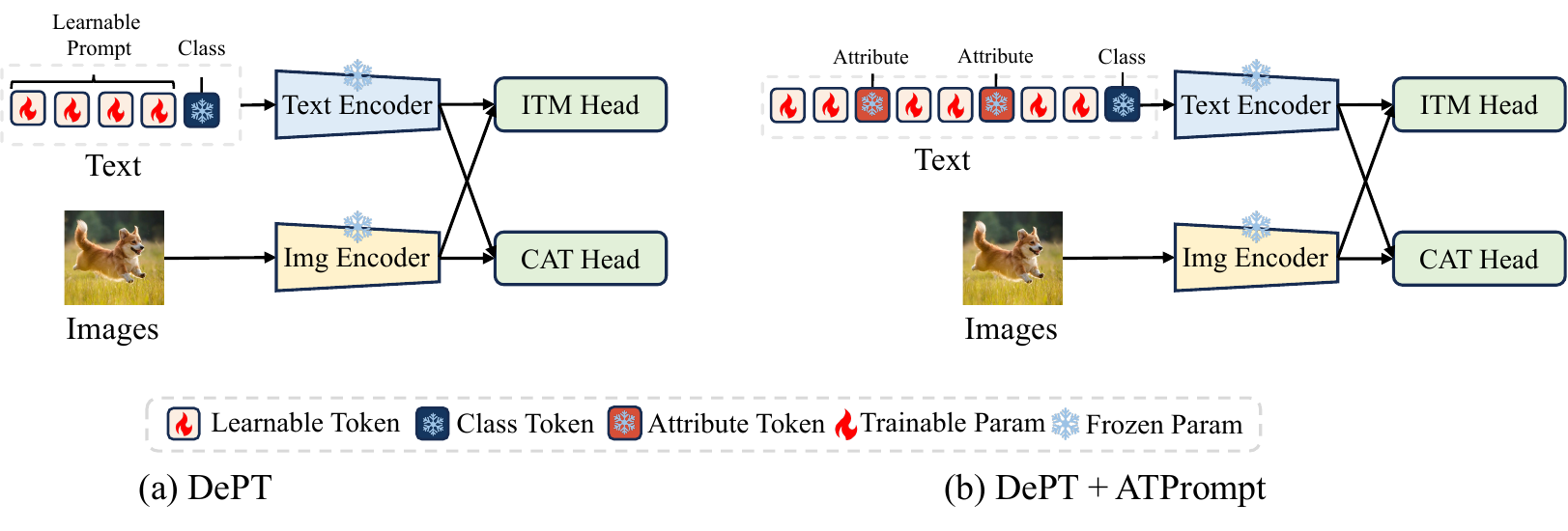}
    \vspace{-10pt}
    \caption{Architectural comparison between DePT and DePT+ATPrompt. }
    \label{fig:comparison_dept}
\end{figure*}

\noindent\textbf{DePT+ATPrompt:} We adopt the baseline's training configuration: a batch size of 32, a learning rate of 0.0035, a balance weight of $\lambda$=0.7, and a duration of 10 epochs. Our primary configuration for DePT+ATPrompt uses a learnable token length of 4. For datasets with lower complexity, namely Caltech-101, OxfordPets, and StanfordCars, we adjust these parameters, setting the soft token length to 2 and the balance weight to 0.6. The architectural differences between the DePT and DePT+ATPrompt models are detailed in Fig.~\ref{fig:comparison_dept}.

% \noindent\textbf{PromptSRC+ATPrompt}: Similar to the baseline method, our approach uses a batch size of 4, an initial learning rate of 0.0025, a text loss weight of 25, and an image loss weight of 10. We adhere to the same training schedule as the baseline method. In the original study, the soft prompt length is specified as 4. In our approach, we set the lengths of both the learnable attribute prompts $a_{m}$ and $b_{m}$, as well as the learnable class soft prompt $M$, to 4.

% In our study, we exclusively remove and later reintroduce deep class soft tokens $[T_{1}]_{i},...,[T_{M}]_{i}$, as explained in our paper. Meanwhile, attribute-related soft tokens $[T_{a_{1}}]_{i}, ..., [T_{b_{m}}]_{i}$, along with hard tokens $[A]_{i}$ and $[B]_{i}$, are retained throughout the forward process. A comprehensive architectural comparison between PromptSRC and PromptSRC+ATPrompt is presented in Fig.~\ref{fig:comparison_promptsrc}.

% \begin{figure*}[!ht]
%     \centering
%     \includegraphics[width=0.85\linewidth]{supp/promptsrc.pdf}
%     \vspace{-10pt}
%     \caption{Architectural comparison between PromptSRC and PromptSRC+ATPrompt. In PromptSRC+ATPrompt, we only drop and reintroduce class soft tokens in the deep layers. The attribute soft and hard tokens are retained throughout the forward process.}
%     \label{fig:comparison_promptsrc}
% \end{figure*}

\section{Additional Experiments}
\subsection{Ablation Study}

\noindent\textbf{Attribute Order.} 
% We have conducted some experiments in the main paper and verified that the order of attributes does not significantly impact model performance. The fluctuation of the results is within a reasonable range. In Tab.~\ref{table:order_of_attribute_more}, we conduct more experiments to verify our observations.
In the main paper, our experiments confirm that the order of attributes does not significantly impact model performance, with results fluctuating within an acceptable range. Here we provide additional experiments in Tab.~\ref{table:order_of_attribute_more} to support this observation.

\begin{table}[ht]
    \centering
    \resizebox{0.74\linewidth}{!}
    {
        \begin{tabular}{cccc}
        \toprule
        Attributes & Base & Novel & HM \\
        \midrule
        (shape, color) & 76.32 & 70.39 & 73.24 \\
        (color, shape) & 76.27 & 70.60 & \textbf{73.33} \\
        \midrule
        (size, habitat) & 76.44 & 70.23 & 73.20 \\
        (habitat, size) & 76.46 & 70.16 & 73.14 \\
        \midrule
        (material, function) & 76.40 & 70.13 & 73.13 \\
        (function, material) & 76.28 & 70.00 & 73.01 \\
        \midrule
        (growth, season) & 76.46 & 70.18 & 73.19 \\
        (season, growth) & 76.40 & 70.21 & 73.17 \\
        \midrule
        (color, size, shape) & 76.27 & 69.95 & 72.97 \\
        (shape, size, color) & 76.32 & 70.19 & 73.13 \\
        \midrule
        (habitat, size, shape) & 76.50 & 70.21 & 73.22 \\
        (habitat, shape, size) & 76.46 & 70.08 & 73.13 \\
        \midrule
        \makecell[c]{Searched Attributes \\ (color, shape)} & 76.27 & 70.60 & \textbf{73.33} \\
        \bottomrule
        \end{tabular}
    }
    \caption{Comparison of different attribute orders on ImageNet. Changes in attribute order will not significantly affect model performance.}
    \label{table:order_of_attribute_more}
    % \vspace{-5pt}
\end{table}

\begin{figure}[!ht]
    \centering
    \includegraphics[width=0.76\linewidth]{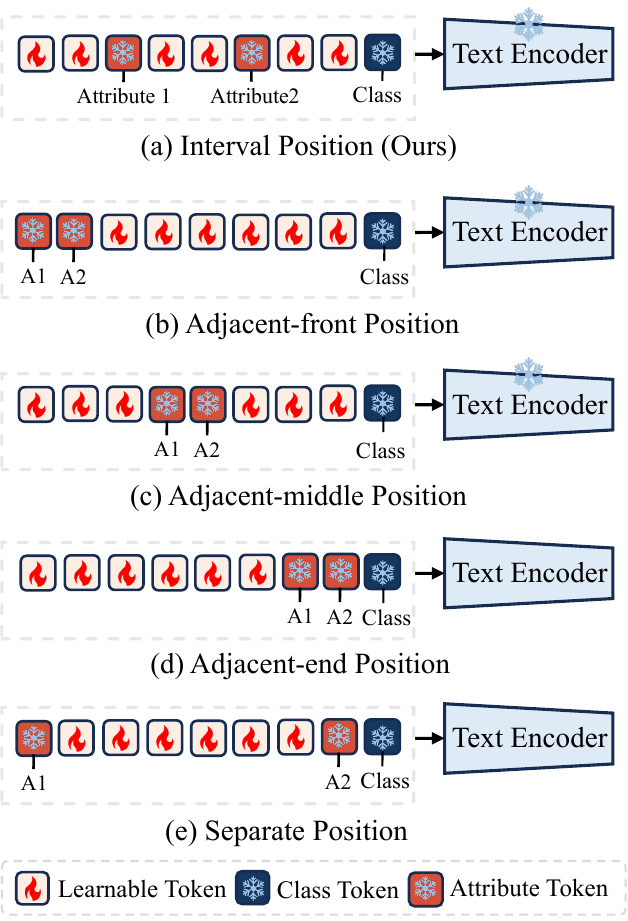}
    % \vspace{-10pt}
    \caption{Comparison of attribute tokens at different positions, taking two attributes as an example.}
    \label{fig:attribute_position}
\end{figure}

\noindent\textbf{Attribute Position.} 
We also investigated the impact of attribute token positioning within the prompt. Fig.~\ref{fig:attribute_position} visualizes the positions tested, and Tab.~\ref{table:attribute_position} presents the results. Our findings show that the "interval" configuration, where attributes are placed between class tokens, yields the best performance.

% In addition to assessing the impact of the class token's position within the textual prompt on performance, we also examine the influence of the attribute token's positioning. Fig.~\ref{fig:attribute_position} and Tab.~\ref{table:attribute_position} present the visualization of attribute positions and the corresponding experimental results, respectively. The results indicate that the interval version achieves the best performance among all variations.

\begin{table}[t]
    \centering
    \resizebox{0.74\linewidth}{!}
    {
        \begin{tabular}{lccc}
        \toprule
        Version         & Base  & Novel & HM    \\
        \midrule
        Baseline~(CoOp) & 76.47 & 67.88 & 71.92 \\
        \midrule
        (a) Interval~(Ours) & 76.27 & 70.60 & 73.33 \\
        (b) Adjacent-front  & 76.39 & 70.22 & 73.18 \\
        (c) Adjacent-middle & 76.46	& 70.11 & 73.15 \\
        (d) Adjacent-end    & 76.34 & 70.31 & 73.20 \\
        (e) Separate        & 76.48 & 70.08 & 73.14 \\
        \bottomrule
        \end{tabular}
    }
    \caption{Performance results of attribute tokens at different positions in ATPrompt on ImageNet. The interval version achieves best results.}
    \label{table:attribute_position}
    % \vspace{-10pt}
\end{table}

\noindent\textbf{Initialization.}
Baseline methods typically initialize soft tokens using the embeddings of the phrase "a photo of a." The inclusion of attribute tokens makes this strategy suboptimal for our method. We instead initialize class soft tokens~($[T_{1}],...,[T_{M}]$) by sampling from a Gaussian distribution with a mean of 0 and a standard deviation of 0.02. As shown in Table~\ref{table:init_compare}, this random initialization provides a superior starting point for training.

% Existing baseline methods initialize the soft tokens using the embeddings of the phrase ``a photo of a" in their official implementations across 11 recognition datasets. However, in our approach, the addition of attribute tokens may render this initialization strategy inappropriate.

% In our method, the class soft tokens $[T_{1}],...,[T_{M}]$ are randomly initialized by sampling from a zero-mean Gaussian distribution with a standard deviation of 0.02. In Tab.~\ref{table:init_compare}, we compare the effects of different initialization methods on model performance. Experimental results show that random initialization provides a superior starting point for training.

\begin{table}[t]
    \centering
    \resizebox{0.74\linewidth}{!}
    {
        \begin{tabular}{cccc}
        \toprule
        Attribute       & Base  & Novel & HM    \\
        \midrule
        ``a photo of a"    & 76.40 & 70.07 & 73.10 \\
        Random Normal Init & 76.27 & 70.60 & 73.33 \\
        \bottomrule
        \end{tabular}
    }
    \caption{Comparison of different initialization ways on ImageNet. Random normal initialization performs better.}
    \label{table:init_compare}
    % \vspace{-10pt}
\end{table}

% \noindent\textbf{Attributes Position.}

% \begin{table}[ht]
%     \centering
%     \resizebox{0.65\linewidth}{!}
%     {
%         \begin{tabular}{cccc}
%         \hline\noalign{\smallskip}
%         Attributes     & Base  & Novel & HM    \\
%         \hline\noalign{\smallskip}
%         Together & 76.32 & 70.39 & 73.24 \\
%         Separate &  \\
%         \hline
%         \end{tabular}
%     }
%     \caption{Comparison of different attribute orders on ImageNet.}
%     \label{table:order_of_attribute}
%     % \vspace{-5pt}
% \end{table}

\section{Discussion}
% \subsection{Intuitive explanation.} 

% As discussed in the main paper, our method utilizes universal attributes as an intermediary to achieve more precise alignment between images and unknown categories. Here is an example to illustrate why this approach is effective. Consider a dataset within the animal domain, where the categories include dog, lion, cat, and bear. We identified the attributes ``shape" and ``color" through LLM and attribute search methods. By utilizing the ``shape" attribute during training, soft prompts can learn general information about the shape, such as the fact that all existing categories possess four legs. When processing downstream data containing a leopard category, these trained soft prompts provide the model with additional information regarding the legs, aiding the model in identifying the leopard from a different perspective.
% dog, lion, cat, bird. attribute: legs.

\textbf{Comparison with Direct LLM Queries.}
Directly querying an LLM for universal attributes presents two challenges: determining the optimal attribute content and identifying the ideal number of attributes. Our experiments suggest that two attributes are often optimal. Therefore, users can bypass our search process by directly prompting the LLM to summarize two universal attributes. This offers a simpler approach, though it may result in a slight performance trade-off.

\noindent\textbf{Why do attributes searched on a source dataset~(ImageNet) generalize well?}
The attributes identified on ImageNet~(e.g., color, shape) are fundamental properties of natural objects. Representations learned under the guidance of these universal attributes are therefore inherently generalizable and transfer effectively to other datasets and classes.

\noindent\textbf{Why does ATPrompt not outperform regularization-based methods in isolation?} ATPrompt is a plug-in module designed to optimize the prompt's structure. In contrast, regularization-based methods are often comprehensive frameworks that employ multiple components (e.g., learnable visual prompts, MLPs) simultaneously. While ATPrompt may not outperform these multi-faceted approaches on its own, its strength lies in its ability to be integrated into other methods, consistently improving their performance beyond previous baselines.

\section{Limitations and Future Works.}
Beyond the limitations discussed in the main paper, we identify the following directions for future research:
(1) While our differentiable search method is efficient, we aim to further enhance the attribute discovery process. A promising direction is to leverage Multimodal Large Language Models (MLLMs), potentially using techniques like Chain-of-Thought (CoT), to better automate the selection of optimal attribute content and quantity.
(2) Our current approach embeds fixed, explicit attributes into the prompt. In the future, we plan to explore a transition to implicit, learnable attributes. This would enable the model to discover optimal attributes in a data-driven manner during training, potentially unlocking further performance gains.

\begin{table*}[ht]
    \centering
    \resizebox{0.73\linewidth}{!}
    {
        \begin{tabular}{ccc}
        \toprule
        Dataset     & Attribute Bases                       & Searched Attributes    \\
        \midrule
        ImageNet-1K & color, size, shape, habitat, behavior & (color, shape) \\
        \midrule
        Caltech-101  & shape, color, material, function, size & (shape,size) \\
        \midrule
        Oxford Pets & loyalty, affection, playfulness, energy, intelligence & (playfulness, energy) \\
        \midrule
        Stanford Cars & design, engine, performance, luxury, color & (luxury) \\
        \midrule
        Flowers-102 & color, flower, habitat, growth, season & (color, habitat, growth) \\
        \midrule
        Food-101    & flavor, texture, origin, ingredients, preparation & (flavor, preparation) \\
        \midrule
        FGVC Aircraft & design, capacity, range, engines, liveries & (design, range) \\
        \midrule
        SUN-397     & architecture, environment, structure, design, function & (function) \\
        \midrule
        DTD        & pattern, texture, color, design, structure & (pattern, color, design) \\
        \midrule
        EuroSAT    & habitat, foliage, infrastructure, terrain, watercourse & (habitat) \\
        \midrule
        UCF-101    & precision, coordination, technique, strength, control & (precision) \\
        \bottomrule
        \end{tabular}
    }
    \caption{Attribute bases and searched results for each dataset.}
    \label{table:attribute_list}
\end{table*}

\begin{table*}[ht]
    \centering
    \resizebox{0.62\linewidth}{!}
    {
        \begin{tabular}{cc}
        \toprule
        Attribute Bases & shape, color, material, function, size \\
        \midrule
        \makecell[c]{Attribute Combinations \\ \& Corresponding Weights} & \makecell[c]{
        (shape), weight: 0.298 \\
        (color), weight: 0.004 \\
        (material), weight: 0.002 \\
        (function), weight: 0.002 \\
        (size), weight: 0.003 \\
        (shape, color), weight: 0.003 \\
        (shape, material), weight: 0.006 \\
        (shape, function), weight: 0.000 \\
        \textbf{(shape, size), weight: 0.565} \\
        (color, material), weight: 0.000 \\
        (color, function), weight: 0.001 \\
        (color, size), weight: 0.005 \\
        (material, function), weight: 0.000 \\
        (material, size), weight: 0.002 \\
        (function, size), weight: 0.002 \\
        (shape, color, material), weight: 0.002 \\
        (shape, color, function), weight: 0.002 \\
        (shape, color, size), weight: 0.000 \\ 
        (shape, material, function), weight: 0.001 \\
        (shape, material, size), weight: 0.085 \\
        (shape, function, size), weight: 0.001 \\
        (color, material, function), weight: 0.001 \\
        (color, material, size), weight: 0.000 \\
        (color, function, size), weight: 0.002 \\
        (material, function, size), weight: 0.001 \\
        (shape, color, material, function), weight: 0.001 \\
        (shape, color, material, size), weight: 0.001 \\
        (shape, color, function, size), weight: 0.001 \\
        (shape, material, function, size), weight: 0.005 \\
        (color, material, function, size), weight: 0.001 \\
        (shape, color, material, function, size), weight: 0.001 \\
        } \\
        \bottomrule
        \end{tabular}
    }
    \caption{Output results after 40 epochs of attribute searching on the Caltech101 dataset.}
    \label{table:combination}
\end{table*}

\end{document}

%% file: arxiv_publish.bbl
\begin{thebibliography}{66}
\providecommand{\natexlab}[1]{#1}
\providecommand{\url}[1]{\texttt{#1}}
\expandafter\ifx\csname urlstyle\endcsname\relax
  \providecommand{\doi}[1]{doi: #1}\else
  \providecommand{\doi}{doi: \begingroup \urlstyle{rm}\Url}\fi

\bibitem[Bahng et~al.(2022)Bahng, Jahanian, Sankaranarayanan, and Isola]{bahng2022exploring}
Hyojin Bahng, Ali Jahanian, Swami Sankaranarayanan, and Phillip Isola.
\newblock Exploring visual prompts for adapting large-scale models.
\newblock \emph{arXiv preprint arXiv:2203.17274}, 2022.

\bibitem[Bar et~al.(2022)Bar, Gandelsman, Darrell, Globerson, and Efros]{bar2022visual}
Amir Bar, Yossi Gandelsman, Trevor Darrell, Amir Globerson, and Alexei Efros.
\newblock Visual prompting via image inpainting.
\newblock \emph{NeurIPS}, 35:\penalty0 25005--25017, 2022.

\bibitem[Bossard et~al.(2014)Bossard, Guillaumin, and Van~Gool]{bossard2014food}
Lukas Bossard, Matthieu Guillaumin, and Luc Van~Gool.
\newblock Food-101--mining discriminative components with random forests.
\newblock In \emph{ECCV}, pages 446--461. Springer, 2014.

\bibitem[Chen et~al.(2023{\natexlab{a}})Chen, Zhang, Han, Chen, Shi, Xu, and Xu]{chen2023vlp}
Fei-Long Chen, Du-Zhen Zhang, Ming-Lun Han, Xiu-Yi Chen, Jing Shi, Shuang Xu, and Bo Xu.
\newblock Vlp: A survey on vision-language pre-training.
\newblock \emph{Machine Intelligence Research}, 20\penalty0 (1):\penalty0 38--56, 2023{\natexlab{a}}.

\bibitem[Chen et~al.(2023{\natexlab{b}})Chen, Jiang, Hu, Tang, Gao, Chen, and Xie]{chen2023ovarnet}
Keyan Chen, Xiaolong Jiang, Yao Hu, Xu Tang, Yan Gao, Jianqi Chen, and Weidi Xie.
\newblock Ovarnet: Towards open-vocabulary object attribute recognition.
\newblock In \emph{CVPR}, pages 23518--23527, 2023{\natexlab{b}}.

\bibitem[Cimpoi et~al.(2014)Cimpoi, Maji, Kokkinos, Mohamed, and Vedaldi]{cimpoi2014describing}
Mircea Cimpoi, Subhransu Maji, Iasonas Kokkinos, Sammy Mohamed, and Andrea Vedaldi.
\newblock Describing textures in the wild.
\newblock In \emph{CVPR}, pages 3606--3613, 2014.

\bibitem[Deng et~al.(2009)Deng, Dong, Socher, Li, Li, and Fei-Fei]{deng2009imagenet}
Jia Deng, Wei Dong, Richard Socher, Li-Jia Li, Kai Li, and Li Fei-Fei.
\newblock Imagenet: A large-scale hierarchical image database.
\newblock In \emph{CVPR}, pages 248--255, 2009.

\bibitem[Ding et~al.(2024)Ding, Li, Miao, and Pfister]{ding2024tree}
Tong Ding, Wanhua Li, Zhongqi Miao, and Hanspeter Pfister.
\newblock Tree of attributes prompt learning for vision-language models.
\newblock \emph{arXiv preprint arXiv:2410.11201}, 2024.

\bibitem[Elsken et~al.(2019)Elsken, Metzen, and Hutter]{elsken2019neural}
Thomas Elsken, Jan~Hendrik Metzen, and Frank Hutter.
\newblock Neural architecture search: A survey.
\newblock \emph{JMLR}, 20\penalty0 (55):\penalty0 1--21, 2019.

\bibitem[Fei-Fei et~al.(2004)Fei-Fei, Fergus, and Perona]{fei2004learning}
Li Fei-Fei, Rob Fergus, and Pietro Perona.
\newblock Learning generative visual models from few training examples: An incremental bayesian approach tested on 101 object categories.
\newblock In \emph{CVPR workshop}, pages 178--178. IEEE, 2004.

\bibitem[Helber et~al.(2019)Helber, Bischke, Dengel, and Borth]{helber2019eurosat}
Patrick Helber, Benjamin Bischke, Andreas Dengel, and Damian Borth.
\newblock Eurosat: A novel dataset and deep learning benchmark for land use and land cover classification.
\newblock \emph{IEEE Journal of Selected Topics in Applied Earth Observations and Remote Sensing}, 12\penalty0 (7):\penalty0 2217--2226, 2019.

\bibitem[Hendrycks et~al.(2021{\natexlab{a}})Hendrycks, Basart, Mu, Kadavath, Wang, Dorundo, Desai, Zhu, Parajuli, Guo, et~al.]{hendrycks2021many}
Dan Hendrycks, Steven Basart, Norman Mu, Saurav Kadavath, Frank Wang, Evan Dorundo, Rahul Desai, Tyler Zhu, Samyak Parajuli, Mike Guo, et~al.
\newblock The many faces of robustness: A critical analysis of out-of-distribution generalization.
\newblock In \emph{ICCV}, pages 8340--8349, 2021{\natexlab{a}}.

\bibitem[Hendrycks et~al.(2021{\natexlab{b}})Hendrycks, Zhao, Basart, Steinhardt, and Song]{hendrycks2021natural}
Dan Hendrycks, Kevin Zhao, Steven Basart, Jacob Steinhardt, and Dawn Song.
\newblock Natural adversarial examples.
\newblock In \emph{CVPR}, pages 15262--15271, 2021{\natexlab{b}}.

\bibitem[Hinton et~al.(2015)Hinton, Vinyals, and Dean]{hinton2015distilling}
Geoffrey Hinton, Oriol Vinyals, and Jeff Dean.
\newblock Distilling the knowledge in a neural network.
\newblock \emph{arXiv preprint arXiv:1503.02531}, 2015.

\bibitem[Huang et~al.(2020)Huang, Li, Wang, Tan, et~al.]{huang2020unfolding}
Yan Huang, Shang Li, Liang Wang, Tieniu Tan, et~al.
\newblock Unfolding the alternating optimization for blind super resolution.
\newblock \emph{NeurIPS}, 33:\penalty0 5632--5643, 2020.

\bibitem[Jia et~al.(2021)Jia, Yang, Xia, Chen, Parekh, Pham, Le, Sung, Li, and Duerig]{jia2021scaling}
Chao Jia, Yinfei Yang, Ye Xia, Yi-Ting Chen, Zarana Parekh, Hieu Pham, Quoc Le, Yun-Hsuan Sung, Zhen Li, and Tom Duerig.
\newblock Scaling up visual and vision-language representation learning with noisy text supervision.
\newblock In \emph{ICML}, pages 4904--4916. PMLR, 2021.

\bibitem[Jia et~al.(2022)Jia, Tang, Chen, Cardie, Belongie, Hariharan, and Lim]{jia2022visual}
Menglin Jia, Luming Tang, Bor-Chun Chen, Claire Cardie, Serge Belongie, Bharath Hariharan, and Ser-Nam Lim.
\newblock Visual prompt tuning.
\newblock In \emph{ECCV}, pages 709--727. Springer, 2022.

\bibitem[Kan et~al.(2023)Kan, Wang, Lu, Zhen, Guan, and Zheng]{kan2023knowledge}
Baoshuo Kan, Teng Wang, Wenpeng Lu, Xiantong Zhen, Weili Guan, and Feng Zheng.
\newblock Knowledge-aware prompt tuning for generalizable vision-language models.
\newblock In \emph{ICCV}, pages 15670--15680, 2023.

\bibitem[Khattak et~al.(2023{\natexlab{a}})Khattak, Rasheed, Maaz, Khan, and Khan]{khattak2023maple}
Muhammad~Uzair Khattak, Hanoona Rasheed, Muhammad Maaz, Salman Khan, and Fahad~Shahbaz Khan.
\newblock Maple: Multi-modal prompt learning.
\newblock In \emph{CVPR}, pages 19113--19122, 2023{\natexlab{a}}.

\bibitem[Khattak et~al.(2023{\natexlab{b}})Khattak, Wasim, Naseer, Khan, Yang, and Khan]{khattak2023self}
Muhammad~Uzair Khattak, Syed~Talal Wasim, Muzammal Naseer, Salman Khan, Ming-Hsuan Yang, and Fahad~Shahbaz Khan.
\newblock Self-regulating prompts: Foundational model adaptation without forgetting.
\newblock In \emph{ICCV}, pages 15190--15200, 2023{\natexlab{b}}.

\bibitem[Kim et~al.(2024)Kim, Kim, and Lee]{kim2024aapl}
Gahyeon Kim, Sohee Kim, and Seokju Lee.
\newblock Aapl: Adding attributes to prompt learning for vision-language models.
\newblock In \emph{CVPR Workshop}, pages 1572--1582, 2024.

\bibitem[Krause et~al.(2013)Krause, Stark, Deng, and Fei-Fei]{krause20133d}
Jonathan Krause, Michael Stark, Jia Deng, and Li Fei-Fei.
\newblock 3d object representations for fine-grained categorization.
\newblock In \emph{ICCV workshop}, pages 554--561, 2013.

\bibitem[Kunananthaseelan et~al.(2024)Kunananthaseelan, Zhang, and Harandi]{kunananthaseelan2024lavip}
Nilakshan Kunananthaseelan, Jing Zhang, and Mehrtash Harandi.
\newblock Lavip: Language-grounded visual prompting.
\newblock In \emph{AAAI}, pages 2840--2848, 2024.

\bibitem[Lee et~al.(2023)Lee, Song, Suh, Choi, Lee, and Kim]{lee2023read}
Dongjun Lee, Seokwon Song, Jihee Suh, Joonmyeong Choi, Sanghyeok Lee, and Hyunwoo~J Kim.
\newblock Read-only prompt optimization for vision-language few-shot learning.
\newblock In \emph{ICCV}, pages 1401--1411, 2023.

\bibitem[Lester et~al.(2021)Lester, Al-Rfou, and Constant]{lester2021power}
Brian Lester, Rami Al-Rfou, and Noah Constant.
\newblock The power of scale for parameter-efficient prompt tuning.
\newblock \emph{arXiv preprint arXiv:2104.08691}, 2021.

\bibitem[Li and Liang(2021)]{li2021prefix}
Xiang~Lisa Li and Percy Liang.
\newblock Prefix-tuning: Optimizing continuous prompts for generation.
\newblock \emph{arXiv preprint arXiv:2101.00190}, 2021.

\bibitem[Li et~al.(2024{\natexlab{a}})Li, Li, Zeng, Wang, Hou, and Cheng]{li2024densevlm}
Yunheng Li, Yuxuan Li, Quansheng Zeng, Wenhai Wang, Qibin Hou, and Ming-Ming Cheng.
\newblock Unbiased region-language alignment for open-vocabulary dense prediction.
\newblock \emph{arXiv preprint arXiv:2412.06244}, 2024{\natexlab{a}}.

\bibitem[Li et~al.(2024{\natexlab{b}})Li, Li, Zeng, Hou, and Cheng]{li2024cascadeclip}
Yunheng Li, Zhong-Yu Li, Quan-Sheng Zeng, Qibin Hou, and Ming-Ming Cheng.
\newblock Cascade-{CLIP}: Cascaded vision-language embeddings alignment for zero-shot semantic segmentation.
\newblock In \emph{ICML}, pages 28243--28258. PMLR, 2024{\natexlab{b}}.

\bibitem[Li et~al.(2023)Li, Li, Yang, Zhao, Song, Luo, Li, and Yang]{li2023curriculum}
Zheng Li, Xiang Li, Lingfeng Yang, Borui Zhao, Renjie Song, Lei Luo, Jun Li, and Jian Yang.
\newblock Curriculum temperature for knowledge distillation.
\newblock In \emph{AAAI}, pages 1504--1512, 2023.

\bibitem[Li et~al.(2024{\natexlab{c}})Li, Li, Fu, Zhang, Wang, Chen, and Yang]{li2024promptkd}
Zheng Li, Xiang Li, Xinyi Fu, Xin Zhang, Weiqiang Wang, Shuo Chen, and Jian Yang.
\newblock Promptkd: Unsupervised prompt distillation for vision-language models.
\newblock In \emph{CVPR}, pages 26617--26626, 2024{\natexlab{c}}.

\bibitem[Li et~al.(2024{\natexlab{d}})Li, Li, Yang, Song, Yang, and Pan]{li2024dual}
Zheng Li, Xiang Li, Lingfeng Yang, Renjie Song, Jian Yang, and Zhigeng Pan.
\newblock Dual teachers for self-knowledge distillation.
\newblock \emph{Pattern Recognition}, 151:\penalty0 110422, 2024{\natexlab{d}}.

\bibitem[Liu et~al.(2018{\natexlab{a}})Liu, Zoph, Neumann, Shlens, Hua, Li, Fei-Fei, Yuille, Huang, and Murphy]{liu2018progressive}
Chenxi Liu, Barret Zoph, Maxim Neumann, Jonathon Shlens, Wei Hua, Li-Jia Li, Li Fei-Fei, Alan Yuille, Jonathan Huang, and Kevin Murphy.
\newblock Progressive neural architecture search.
\newblock In \emph{ECCV}, pages 19--34, 2018{\natexlab{a}}.

\bibitem[Liu et~al.(2018{\natexlab{b}})Liu, Simonyan, and Yang]{liu2018darts}
Hanxiao Liu, Karen Simonyan, and Yiming Yang.
\newblock Darts: Differentiable architecture search.
\newblock \emph{arXiv preprint arXiv:1806.09055}, 2018{\natexlab{b}}.

\bibitem[Lu et~al.(2019)Lu, Batra, Parikh, and Lee]{lu2019vilbert}
Jiasen Lu, Dhruv Batra, Devi Parikh, and Stefan Lee.
\newblock Vilbert: Pretraining task-agnostic visiolinguistic representations for vision-and-language tasks.
\newblock \emph{NeurIPS}, 32, 2019.

\bibitem[Ma et~al.(2024)Ma, Li, Yang, Patashnik, Lischinski, Cohen-Or, and Huang]{ma2024clip}
Hao Ma, Ming Li, Jingyuan Yang, Or Patashnik, Dani Lischinski, Daniel Cohen-Or, and Hui Huang.
\newblock Clip-flow: Decoding images encoded in clip space.
\newblock \emph{CVMJ}, 10\penalty0 (6):\penalty0 1157--1168, 2024.

\bibitem[Maji et~al.(2013)Maji, Rahtu, Kannala, Blaschko, and Vedaldi]{maji2013fine}
Subhransu Maji, Esa Rahtu, Juho Kannala, Matthew Blaschko, and Andrea Vedaldi.
\newblock Fine-grained visual classification of aircraft.
\newblock \emph{arXiv preprint arXiv:1306.5151}, 2013.

\bibitem[Menon and Vondrick(2022)]{menon2022visual}
Sachit Menon and Carl Vondrick.
\newblock Visual classification via description from large language models.
\newblock \emph{arXiv preprint arXiv:2210.07183}, 2022.

\bibitem[Nilsback and Zisserman(2008)]{nilsback2008automated}
Maria-Elena Nilsback and Andrew Zisserman.
\newblock Automated flower classification over a large number of classes.
\newblock In \emph{2008 Sixth Indian conference on computer vision, graphics \& image processing}, pages 722--729. IEEE, 2008.

\bibitem[Parkhi et~al.(2012)Parkhi, Vedaldi, Zisserman, and Jawahar]{parkhi2012cats}
Omkar~M Parkhi, Andrea Vedaldi, Andrew Zisserman, and CV Jawahar.
\newblock Cats and dogs.
\newblock In \emph{CVPR}, pages 3498--3505. IEEE, 2012.

\bibitem[Paszke et~al.(2019)Paszke, Gross, Massa, Lerer, Bradbury, Chanan, Killeen, Lin, Gimelshein, Antiga, et~al.]{paszke2019pytorch}
Adam Paszke, Sam Gross, Francisco Massa, Adam Lerer, James Bradbury, Gregory Chanan, Trevor Killeen, Zeming Lin, Natalia Gimelshein, Luca Antiga, et~al.
\newblock Pytorch: An imperative style, high-performance deep learning library.
\newblock In \emph{NeurIPS}, pages 8026--8037, 2019.

\bibitem[Pham et~al.(2018)Pham, Guan, Zoph, Le, and Dean]{pham2018efficient}
Hieu Pham, Melody Guan, Barret Zoph, Quoc Le, and Jeff Dean.
\newblock Efficient neural architecture search via parameters sharing.
\newblock In \emph{ICML}, pages 4095--4104. PMLR, 2018.

\bibitem[Radford et~al.(2021)Radford, Kim, Hallacy, Ramesh, Goh, Agarwal, Sastry, Askell, Mishkin, Clark, et~al.]{radford2021learning}
Alec Radford, Jong~Wook Kim, Chris Hallacy, Aditya Ramesh, Gabriel Goh, Sandhini Agarwal, Girish Sastry, Amanda Askell, Pamela Mishkin, Jack Clark, et~al.
\newblock Learning transferable visual models from natural language supervision.
\newblock In \emph{ICML}, pages 8748--8763. PMLR, 2021.

\bibitem[Recht et~al.(2019)Recht, Roelofs, Schmidt, and Shankar]{recht2019imagenet}
Benjamin Recht, Rebecca Roelofs, Ludwig Schmidt, and Vaishaal Shankar.
\newblock Do imagenet classifiers generalize to imagenet?
\newblock In \emph{ICML}, pages 5389--5400. PMLR, 2019.

\bibitem[Roy and Etemad(2023)]{roy2023consistency}
Shuvendu Roy and Ali Etemad.
\newblock Consistency-guided prompt learning for vision-language models.
\newblock \emph{arXiv preprint arXiv:2306.01195}, 2023.

\bibitem[Soomro et~al.(2012)Soomro, Zamir, and Shah]{soomro2012ucf101}
Khurram Soomro, Amir~Roshan Zamir, and Mubarak Shah.
\newblock Ucf101: A dataset of 101 human actions classes from videos in the wild.
\newblock \emph{arXiv preprint arXiv:1212.0402}, 2012.

\bibitem[Sun et~al.(2024)Sun, Fang, Wu, Zhang, Zang, Kong, Xiong, Lin, and Wang]{sun2024alpha}
Zeyi Sun, Ye Fang, Tong Wu, Pan Zhang, Yuhang Zang, Shu Kong, Yuanjun Xiong, Dahua Lin, and Jiaqi Wang.
\newblock Alpha-clip: A clip model focusing on wherever you want.
\newblock In \emph{CVPR}, pages 13019--13029, 2024.

\bibitem[Tan and Bansal(2019)]{tan2019lxmert}
Hao Tan and Mohit Bansal.
\newblock Lxmert: Learning cross-modality encoder representations from transformers.
\newblock \emph{arXiv preprint arXiv:1908.07490}, 2019.

\bibitem[Tan et~al.(2019)Tan, Chen, Pang, Vasudevan, Sandler, Howard, and Le]{tan2019mnasnet}
Mingxing Tan, Bo Chen, Ruoming Pang, Vijay Vasudevan, Mark Sandler, Andrew Howard, and Quoc~V Le.
\newblock Mnasnet: Platform-aware neural architecture search for mobile.
\newblock In \emph{CVPR}, pages 2820--2828, 2019.

\bibitem[Tian et~al.(2024)Tian, Zou, Yang, and Zhang]{tian2024argue}
Xinyu Tian, Shu Zou, Zhaoyuan Yang, and Jing Zhang.
\newblock Argue: Attribute-guided prompt tuning for vision-language models.
\newblock In \emph{CVPR}, pages 28578--28587, 2024.

\bibitem[Wang et~al.(2019)Wang, Ge, Lipton, and Xing]{wang2019learning}
Haohan Wang, Songwei Ge, Zachary Lipton, and Eric~P Xing.
\newblock Learning robust global representations by penalizing local predictive power.
\newblock \emph{NeurIPS}, 32, 2019.

\bibitem[Wang et~al.(2024)Wang, Jiang, Cheng, Li, and Zhao]{wang2024learning}
Yubin Wang, Xinyang Jiang, De Cheng, Dongsheng Li, and Cairong Zhao.
\newblock Learning hierarchical prompt with structured linguistic knowledge for vision-language models.
\newblock In \emph{AAAI}, pages 5749--5757, 2024.

\bibitem[Wei et~al.(2022)Wei, Wang, Schuurmans, Bosma, Xia, Chi, Le, Zhou, et~al.]{wei2022chain}
Jason Wei, Xuezhi Wang, Dale Schuurmans, Maarten Bosma, Fei Xia, Ed Chi, Quoc~V Le, Denny Zhou, et~al.
\newblock Chain-of-thought prompting elicits reasoning in large language models.
\newblock \emph{NeurIPS}, 35:\penalty0 24824--24837, 2022.

\bibitem[Wu et~al.(2024)Wu, Zhang, Li, Chen, Liang, Yang, and Li]{wu2024cascade}
Ge Wu, Xin Zhang, Zheng Li, Zhaowei Chen, Jiajun Liang, Jian Yang, and Xiang Li.
\newblock Cascade prompt learning for vision-language model adaptation.
\newblock In \emph{ECCV}, 2024.

\bibitem[Wu et~al.(2025)Wu, Zhang, Zeng, Gao, Song, and Shen]{wu2025skip}
Shihan Wu, Ji Zhang, Pengpeng Zeng, Lianli Gao, Jingkuan Song, and Heng~Tao Shen.
\newblock Skip tuning: Pre-trained vision-language models are effective and efficient adapters themselves.
\newblock In \emph{CVPR}, pages 14723--14732, 2025.

\bibitem[Xiao et~al.(2010)Xiao, Hays, Ehinger, Oliva, and Torralba]{xiao2010sun}
Jianxiong Xiao, James Hays, Krista~A Ehinger, Aude Oliva, and Antonio Torralba.
\newblock Sun database: Large-scale scene recognition from abbey to zoo.
\newblock In \emph{CVPR}, pages 3485--3492. IEEE, 2010.

\bibitem[Yang et~al.(2023{\natexlab{a}})Yang, An, Huang, Bi, Yu, Yang, and Xu]{yang2023clip}
Chuanguang Yang, Zhulin An, Libo Huang, Junyu Bi, Xinqiang Yu, Han Yang, and Yongjun Xu.
\newblock Clip-kd: An empirical study of distilling clip models.
\newblock \emph{arXiv preprint arXiv:2307.12732}, 2023{\natexlab{a}}.

\bibitem[Yang et~al.(2023{\natexlab{b}})Yang, An, Zhou, Zhuang, Xu, and Zhang]{yang2023online}
Chuanguang Yang, Zhulin An, Helong Zhou, Fuzhen Zhuang, Yongjun Xu, and Qian Zhang.
\newblock Online knowledge distillation via mutual contrastive learning for visual recognition.
\newblock \emph{IEEE Transactions on Pattern Analysis and Machine Intelligence}, 45\penalty0 (8):\penalty0 10212--10227, 2023{\natexlab{b}}.

\bibitem[Yao et~al.(2023)Yao, Zhang, and Xu]{yao2023visual}
Hantao Yao, Rui Zhang, and Changsheng Xu.
\newblock Visual-language prompt tuning with knowledge-guided context optimization.
\newblock In \emph{CVPR}, pages 6757--6767, 2023.

\bibitem[Zhai et~al.(2024)Zhai, Zeng, Huang, Qin, Jin, and Cao]{zhai2024multi}
Yajing Zhai, Yawen Zeng, Zhiyong Huang, Zheng Qin, Xin Jin, and Da Cao.
\newblock Multi-prompts learning with cross-modal alignment for attribute-based person re-identification.
\newblock In \emph{AAAI}, pages 6979--6987, 2024.

\bibitem[Zhang et~al.(2024)Zhang, Wu, Gao, Shen, and Song]{zhang2024dept}
Ji Zhang, Shihan Wu, Lianli Gao, Heng~Tao Shen, and Jingkuan Song.
\newblock Dept: Decoupled prompt tuning.
\newblock In \emph{CVPR}, pages 12924--12933, 2024.

\bibitem[Zhang et~al.(2025)Zhang, Yu, Zhao, Fan, and Xiao]{zhang2025comprompter}
Xiaoqin Zhang, Zhenni Yu, Li Zhao, Deng-Ping Fan, and Guobao Xiao.
\newblock Comprompter: reconceptualized segment anything model with multiprompt network for camouflaged object detection.
\newblock \emph{Science China Information Sciences}, 68\penalty0 (1):\penalty0 112104, 2025.

\bibitem[Zhou et~al.(2022{\natexlab{a}})Zhou, Sch{\"a}rli, Hou, Wei, Scales, Wang, Schuurmans, Cui, Bousquet, Le, et~al.]{zhou2022least}
Denny Zhou, Nathanael Sch{\"a}rli, Le Hou, Jason Wei, Nathan Scales, Xuezhi Wang, Dale Schuurmans, Claire Cui, Olivier Bousquet, Quoc Le, et~al.
\newblock Least-to-most prompting enables complex reasoning in large language models.
\newblock \emph{arXiv preprint arXiv:2205.10625}, 2022{\natexlab{a}}.

\bibitem[Zhou et~al.(2022{\natexlab{b}})Zhou, Yang, Loy, and Liu]{zhou2022conditional}
Kaiyang Zhou, Jingkang Yang, Chen~Change Loy, and Ziwei Liu.
\newblock Conditional prompt learning for vision-language models.
\newblock In \emph{CVPR}, pages 16816--16825, 2022{\natexlab{b}}.

\bibitem[Zhou et~al.(2022{\natexlab{c}})Zhou, Yang, Loy, and Liu]{zhou2022learning}
Kaiyang Zhou, Jingkang Yang, Chen~Change Loy, and Ziwei Liu.
\newblock Learning to prompt for vision-language models.
\newblock \emph{IJCV}, 130\penalty0 (9):\penalty0 2337--2348, 2022{\natexlab{c}}.

\bibitem[Zhu et~al.(2023)Zhu, Niu, Han, Wu, and Zhang]{zhu2023prompt}
Beier Zhu, Yulei Niu, Yucheng Han, Yue Wu, and Hanwang Zhang.
\newblock Prompt-aligned gradient for prompt tuning.
\newblock In \emph{ICCV}, pages 15659--15669, 2023.

\bibitem[Zoph et~al.(2018)Zoph, Vasudevan, Shlens, and Le]{zoph2018learning}
Barret Zoph, Vijay Vasudevan, Jonathon Shlens, and Quoc~V Le.
\newblock Learning transferable architectures for scalable image recognition.
\newblock In \emph{CVPR}, pages 8697--8710, 2018.

\end{thebibliography}
